\documentclass[onecolumn]{article}
\usepackage{graphicx}
\usepackage{amsmath}
\usepackage{amssymb}
\usepackage{times}
\usepackage{verbatim}
\usepackage{algorithm}
\usepackage{algorithmic}

\def\BibTeX{{\rm B\kern-.05em{\sc i\kern-.025em b}\kern-.08em
    T\kern-.1667em\lower.7ex\hbox{E}\kern-.125emX}}

\begin{document}

\title{Local Optima Networks \\ of NK Landscapes with Neutrality}

\author{S\'ebastien Verel, Gabriela Ochoa, Marco Tomassini\thanks{
        S\'ebastien Verel is with INRIA Lille - Nord Europe, and University of Nice Sophia-Antipolis~/~CNRS, Nice, France. E-mail: {\tt verel@i3s.unice.fr}.
        Gabriela Ochoa is with The Automated Scheduling, Optimisation and Planning (ASAP) Group, School of Computer Science, University of Nottingham, Nottingham, UK. E-mail: {\tt gxo@cs.nott.ac.uk}.
        Marco Tomassini is with the Information Systems Department, HEC, University of Lausanne, Switzerland. E-mail: {\tt Marco.Tomassini@unil.ch}}}

\date{ }

\markboth{Verel et al.: Local Optima Networks of NK Landscapes with Neutrality}
{Verel et al.: Local Optima Networks of NK Landscapes with Neutrality}

\maketitle

\begin{abstract}
In previous work, we have introduced a network-based model that abstracts many details of the
underlying landscape and compresses the landscape information into a weighted, oriented graph which we call the \textit{local optima network}. The vertices of this graph are the local optima of the given fitness landscape, while the arcs are transition probabilities between local optima basins. Here, we extend this formalism to neutral fitness landscapes, which are  common in difficult combinatorial search spaces. By using two known neutral variants of the $NK$ family (i.e. $NK_p$ and $NK_q$) in which the amount of neutrality can be tuned by a parameter, we show that our new definitions of the optima networks and the associated basins are consistent with the previous definitions for the non-neutral case. Moreover, our empirical study and statistical analysis show that the features of neutral landscapes interpolate smoothly between landscapes with maximum neutrality and non-neutral ones. We found some unknown structural differences between the two studied families of neutral landscapes. But overall, the network features studied confirmed that neutrality, in landscapes with percolating neutral networks, may enhance heuristic search. Our current methodology requires the exhaustive enumeration of the underlying search space. Therefore, sampling techniques should be developed before this analysis can have  practical implications. We argue, however, that the proposed model offers a new perspective into the problem difficulty of combinatorial optimization problems and may inspire the design of more effective search heuristics.
\end{abstract}


\section{Introduction}
\label{intro}

Studying the distribution of local optima in a search space is of utmost importance for understanding the search difficulty of the corresponding landscape. This understanding may eventually be exploited when designing efficient search algorithms. For example, it has been observed in many combinatorial landscapes that local optima are not randomly distributed, rather they tend to be clustered in a "central massif" (or "big valley" if we are minimizing). This globally convex landscape structure has been observed in the $NK$ family of landscapes \cite{Kauffman1987,kauffman93}, and in many combinatorial optimization problems, such as the traveling salesman problem \cite{boese94}, graph bipartitioning~\cite{merz98}, and flowshop scheduling \cite{reeves99}. Algorithms that exploit this global structure have, in consequence, been proposed~\cite{boese94,reeves99}.

\begin{figure} [!ht]
\begin{center}
\includegraphics[width=0.3\textwidth]{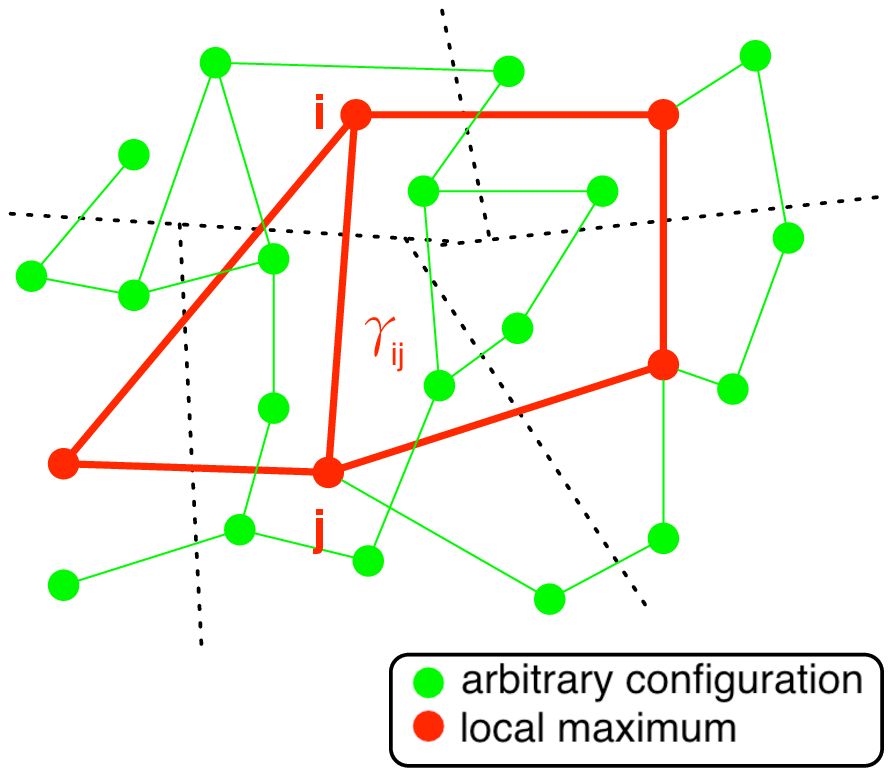} \\
\includegraphics[width=0.3\textwidth]{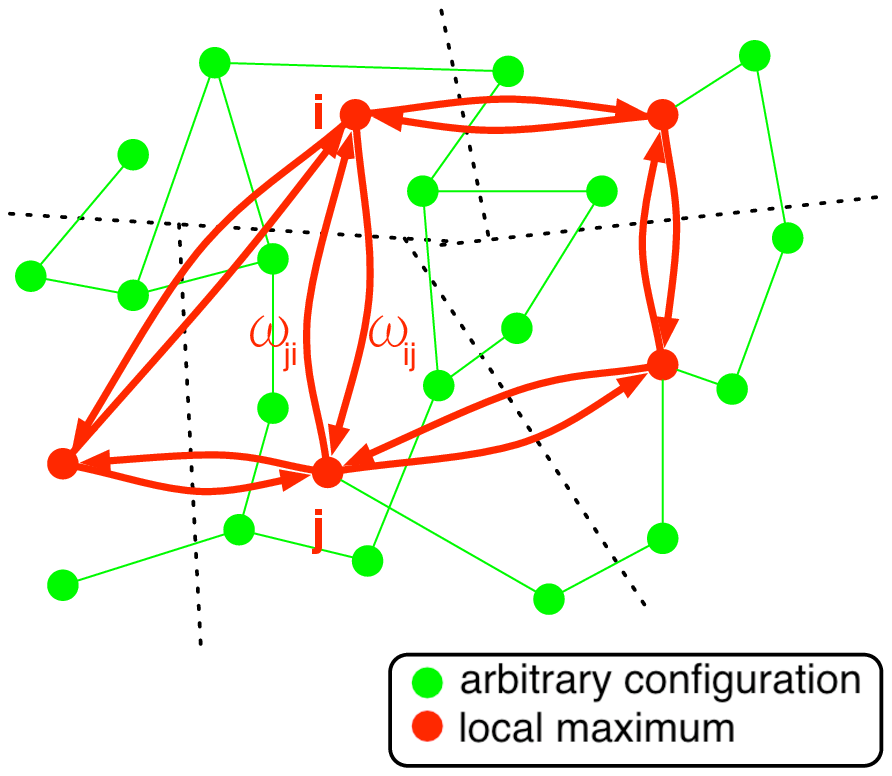} \\
\vspace{-0.3cm} \caption{A diagram of the {\em local optima} or {\em basin adjacency } networks. The dark nodes correspond to the local optima in the landscape, whereas the edges  represent the notion of adjacency among basins. Dashed lines separate the basins. Two alternative definitions of edges are sketched as undirected (top plot) and directed  weighted arcs (bottom). \label{fig:networks}}
\end{center}
\end{figure}

Combinatorial landscapes can be seen as a graph whose vertices are the possible configurations. If two configurations can be transformed into each other by a suitable operator move, then we can trace an edge between them. The resulting graph, with an indication of the fitness at each vertex, is a representation of the given problem fitness landscape. A useful simplification of the graphs for the energy landscapes of atomic clusters was introduced in \cite{doye02,doye05}. The idea consists of taking as vertices of the graph not all the possible configurations, but only those that correspond to energy minima. For atomic clusters these are well-known, at least for relatively small assemblages. Two minima are considered connected, and thus an edge is traced between them, if the energy barrier separating them is sufficiently low. In this case there is a transition state, meaning that the system can jump from one minimum to the other by thermal fluctuations going through a saddle point in the energy hyper-surface. The values of these activation energies are mostly known experimentally or can be determined by simulation. In this way, a network can be built which is called the "inherent structure" or "inherent network" in~\cite{doye02}.\\
In~\cite{gecco08,alife08,pre09}, we proposed a network characterization of combinatorial fitness landscapes by adapting the notion of inherent networks described above. We used the well-known family of $NK$ landscapes as an example. In our case, the inherent network was the graph where the vertices are all the local maxima, obtained exhaustively by running a best-improvement (steepest-ascent) local search algorithm  from every configuration of the search space. The edges accounted for the notion of adjacency between basins. In our work we call this graph the \textit{local optima network} or since it also represents the interaction between the landscape's basin the \textit{basin adjacency network}. We proposed two alternative definitions of edges. In the first definition \cite{gecco08}, two maxima $i$ and $j$ were connected (with an undirected edge without weight), if there exists at least one pair of directly connected solutions $s_i$ and $s_j$, one in each basin of attraction ($b_i$ and $b_j$) (Fig.~\ref{fig:networks}, top). The second, more accurate definition, associated weights to the edges that account for the transition probabilities between the basins of attraction of the local optima (Fig.~\ref{fig:networks}, bottom). More details on the relevant algorithms and formal definitions are given in section \ref{defs}. This characterization of landscapes as networks has brought new insights into the global structure of the landscapes studied, particularly into the distribution of their local optima. Therefore, the application of these techniques to more realistic and complex landscapes, is a research direction worth exploring.

The fitness landscape metaphor \cite{Wright32} has been a standard tool for visualizing biological evolution and speciation. It has also been useful for studying  the dynamics of evolutionary and heuristic search algorithms applied to optimization and design problems. Traditionally, fitness landscapes are often depicted as `rugged' surfaces with many local `peaks' of different heights flanked by `valleys' of different depth \cite{Kauffman1987,kauffman93}. This view is now acknowledged to be only part of the story. In both natural and artificial systems a picture is emerging of populations engaged not in hill-climbing, but rather drifting along connected networks of genotypes of equal (or quasi equal) fitness, with sporadic jumps between these so called  $neutral$ networks. The importance of {\em selective neutrality} as a significant factor in evolution was stressed by Kimura \cite{Kimura1983} in the context of evolutionary theory, and by Eigen et al. \cite{Eigen1988} in the context of molecular biology.   Interest in selective neutrality was re-gained in the 90s by the identification of neutral networks in models for bio-polymer sequence to structure mappings \cite{Fontana1993,Gruener1996,Gruener1996a,Huynen1996,Reidys1997,schuster1,schuster2}.  It has also been observed that the huge dimensionality of biologically interesting fitness landscapes, considering the redundancy in the genotype-fitness map, brings naturally the existence of neutral and nearly neutral networks \cite{Gavrilets2004}. In this context, the metaphor of `holey adaptive landscapes' has been put forward as an alternative to the conventionally view of rugged adaptive landscapes, to model macro-evolution and speciation in nature \cite{Gavrilets2004,Gavrilets1997,Gavrilets1997a}. The relevance and benefits of neutrality for the robustness and evolvability in living systems has been recently discussed in  \cite{Wagner2005}.

There is growing evidence that such large-scale neutrality is also present in artificial landscapes. Not only in combinatorial fitness landscapes such as randomly generated SAT instances~\cite{frank-et-al97},  cellular automata rules~\cite{verel07} and many others,  but also  in complex real-world design and engineering applications such as evolutionary robotics~\cite{husbands,Smith2002}, evolvable hardware~\cite{HarveyT96,Thompson1999,Vassilev2000},  genetic programming~\cite{Banzhaf1994,ebner,yu01,Miller2006} and grammatical evolution \cite{Wilson2009}.

Not only the structure of interesting natural and artificial landscapes, as discussed above, is different from the conventional view of rugged landscapes; the evidence also suggests that the {\em dynamics} of evolutionary (or more generally search) processes on fitness landscapes with neutrality are qualitatively very different from the dynamics on rugged landscapes \cite{Huynen1996,HarveyT96,Nimwegen1997,Barnett:97,barnett98,barnett01,Reidys2000,Reidys2001}. As a consequence,  techniques for effective evolutionary search on landscapes with neutrality may be quite different from more traditional approaches to evolutionary search~\cite{barnett01,VEREL:2004:HAL-00160035:1}.

In this paper, we apply our previous network definitions and analysis of combinatorial search spaces to landscapes with selective neutrality. In particular, it is our intention to investigate whether our graph-based approach is still adequate when neutrality is present. This is apparently simple but, in reality, requires a careful redefinition of the concept of a basin of attraction. The new notions will be presented in the next section.  We also study how neutrality affects the landscape graph structure and statistics, and discuss the implications for the dynamic of heuristic search on these landscapes. Following our previous work on $NK$ landscapes~\cite{gecco08,alife08,pre09}, we selected two extensions of the $NK$ family as example landscapes with synthetic neutrality, namely: the $NK_p$ (`probabilistic' $NK$)~\cite{barnett98}, and $NK_q$ (`quantized' $NK$)~\cite{newman98} families. The  $NK_p$  landscape introduces neutrality  by  setting  a  certain  proportion  $p$  of  the  entries  in  a  genotype's fitness tables to 0; whilst the  $NK_q$ landscape does so by transforming the genotype fitness entries from real numbers to integer values (in the range [0, q)). These landscapes posses two statistical features: fitness correlation and selective neutrality, which are relevant to combinatorial optimization.

The paper begins by describing in more detail the neutral families of landscapes under study (section \ref{neutralNK}). Thereafter, section~\ref{defs} includes the relevant definitions and algorithms used. The empirical network analysis of our selected neutral landscape instances is presented next (section~\ref{analysis}), followed by a summary and discussion (section~\ref{discussion}) and our conclusions and ideas for future work (section~\ref{conclusions}).

\section{$NK$ landscapes with neutrality}
\label{neutralNK}

The $NK$ family of landscapes \cite{kauffman93} is a problem-independent model for constructing multimodal landscapes that can gradually be tuned from smooth to rugged. In the model, $N$ refers to the number of (binary) genes in the genotype (i.e. the string length) and $K$ to the number of genes that influence a particular gene (the epistatic interactions). By increasing the value of $K$ from 0 to $N-1$, $NK$ landscapes can be tuned from smooth to rugged.

The fitness function of a $NK$-landscape $f_{NK}: \lbrace 0, 1 \rbrace^{N} \rightarrow [0,1)$ is defined on binary strings with $N$ bits. An `atom' with fixed epistasis level is represented by a fitness component $f_i: \lbrace 0, 1 \rbrace^{K+1} \rightarrow [0,1)$ associated to each bit $i$.
Its value depends on the allele at bit $i$ and also on the alleles at the  $K$ other epistatic positions.
($K$ must fall between $0$ and $N - 1$).
The fitness $f_{NK}(s)$ of $s \in \lbrace 0, 1 \rbrace^{N}$ is the average of the values of the $N$ fitness components $f_i$: \label{defNK}
$$ f_{NK}(s) = \frac{1}{N} \sum_{i=1}^{N} f_i(s_i, s_{i_1}, \ldots, s_{i_K})
$$ where $\lbrace i_1, \ldots, i_{K} \rbrace \subset \lbrace 1, \ldots, i - 1, i + 1, \ldots, N \rbrace$.
Several ways have been proposed to choose the $K$ other bits
from $N$ bits in the bit string. Two possibilities are mainly used: adjacent and random neighborhoods.
With an adjacent neighborhood, the $K$ bits nearest to the bit $i$ are chosen (the genotype is taken to have periodic boundaries). With a random neighborhood,
the $K$ bits are chosen randomly on the bit string.
Each fitness component $f_i$ is specified by extension,
\textit{i.e.} a number $y^i_{s_i, s_{i_1}, \ldots, s_{i_K}}$ from $[0, 1)$  is associated with each element
$(s_i, s_{i_1}, \ldots, s_{i_K})$ from $\lbrace 0, 1 \rbrace^{K+1}$.
Those numbers are uniformly distributed in the range $[0, 1)$.

The two variants of $NK$ landscapes are representative of the way to obtain neutrality in additive fitness landscapes. Indeed, for the two families, the fitness value of a solution is computed as a sum. Modifying a term in the sum would alter the probability to get the same fitness value.

The {\it $NK_p$ landscapes} have been introduced by Barnett \cite{barnett98}.
In this variant, one term of the sum is null with probability $p$. Formally, the fitness components are modified and tuned by the parameter $p \in [0,1]$ which controls the neutrality of the landscape. The fitness component $y^i_{s_i, s_{i_1}, \ldots, s_{i_K}}$ is null with probability $p$, \textit{i.e.} $P(y^i_{s_i, s_{i_1}, \ldots, s_{i_K}} = 0) = p$.
The probability that two neighboring solutions have the same fitness value increases with the parameter $p$.

The \textit{$NK_q$ landscapes} have been introduced by Newman {\it et al}~\cite{newman98}.
For these landscapes, the terms of the sum are integer numbers between $0$ and $q-1$.
Thus, when some terms are modified, it is possible to get the same sum.
Formally, as for $NK_p$ landscapes, the fitness components are defined with a parameter $q$ which tunes the neutrality. Parameter $q$ is an integer number above or equal to $2$.
Each $y^i_{s_i, s_{i_1}, \ldots, s_{i_K}}$ is one of the fractions $\frac{k}{q}$
where $k$ is an integer number randomly chosen in $[0, q-1]$.

Neutrality is maximal when $q$ is equal to $2$, and decreases when $q$ increases.  This family of landscapes was shown to model the properties of neutral evolution of molecular species~\cite{newman98}.

\section{Definitions and Algorithms}
\label{defs}

We include the relevant definitions and algorithms to obtain the local optima network in landscapes with neutrality. For completeness, we also include some relevant definitions that apply to non-neutral landscapes \cite{alife08,pre09}.

\textbf{Fitness landscape:}

A landscape is a triplet $(S, V, f)$ where $S$ is a set of admissible solutions i.e. a search space, $V : S \longrightarrow 2^{|S|}$, a neighborhood structure, is a function that assigns to every $s \in
S$ a set of neighbors $V(s)$, and $f : S \longrightarrow R$ is a fitness function that can be pictured as the \textit{height} of the corresponding solutions.

In our study, the search space is composed of binary strings of length $N$, therefore its size is $2^N$. The neighborhood is defined by the minimum possible move on a binary search space, that is, the 1-move or bit-flip operation. In consequence, for any given string $s$ of length $N$, the neighborhood size is $|V(s)| = N$.

\textbf{Neutral neighbor:}
A neutral neighbor of $s$ is a neighbor configuration $x$ with the same fitness $f(s)$.
$$V_n(s) = \{ x \in V(s) ~|~ f(x) = f(s) \}$$

The neutral degree of a solution is the number of its neutral neighbors.

A fitness landscape is neutral if there are many solutions with high neutral degree. The landscape is then composed of- several sub-graphs of configurations with the same fitness value. Sometimes, another definition of neutral neighbor is used in which the fitness values are allowed to differ by a small amount. Here we stick to the strict definitions given above.

\textbf{Neutral network:}
A neutral network, denoted as $NN$, is a connected sub-graph whose vertices are configurations with the same fitness value. Two vertices in a $NN$ are connected if they are neutral neighbors.

With the bit-flip mutation operator, for all solutions $x$ and $y$, if $x \in V(y)$ then $y \in V(x)$.
So in this case, the neutral networks are the equivalent classes of the relation $R(x,y)$ iff ($x \in V(y)$ and $f(x)=f(y)$)\footnote{Our definition of neutrality is strict. It also possible to define a concept of \textit{quasi-neutrality}~\cite{verel07} but we do not use it in this work.}.

We denote the neutral network of a configuration $s$ by $NN(s)$.

\subsection{Definition of basins of attraction}

In this section, we define the notion of a basin of attraction for landscapes with neutrality. The analogous notion for non-neutral landscapes has been given in~\cite{pre09}.

First let us define the standard notion of a local optimum, and its extension for landscapes with neutral networks.

\textbf{Local optimum:}
A local optimum, which is taken to be a maximum here, is a solution $s^{*}$ such that $\forall s \in V(s)$, $f(s) \leq f(s^{*})$.\\

Notice that the inequality is not strict, in order to allow the treatment of the neutral landscape case.\\

\textbf{Local optimum neutral network (LONN):}
A neutral network is a local optimum if all the configurations of the neutral network are local optima.\\

To extract the basins of attraction of the local optima neutral networks,  the "Stochastic Hill Climbing" algorithm is used. In this algorithm (illustrated below) one neighbour solution with maximum fitness is randomly chosen, and solutions with equal or improved fitness are accepted.
\begin{algorithm}
\caption{Stochastic Hill Climbing}
\label{algoHC}
\begin{algorithmic}
\STATE Choose initial solution $s \in \cal S$
\REPEAT
	\STATE randomly choose  $s^{'}$ from $ \{ z \in V(s) | f(z) = max \{ f(x) | x \in V(s) \} \}$
        \IF{$f(s) \leq f(s^{'})$}
        	\STATE $s \leftarrow s^{'}$
	\ENDIF
\UNTIL{$s$ is in a LONN}
\end{algorithmic}
\end{algorithm}

Let us denote by $h$, the stochastic operator which associates to each solution $s$,
the solution obtained after applying the Stochastic Hill Climbing algorithm for a sufficiently large number of iterations to converge to a solution in a LONN.

The size of the landscape is finite, so we can denote by $NN_1$, $NN_2$, $NN_3 \ldots, NN_n$, the local optima neutral networks. These LONNs are the vertices of the \textit{local optima network} in the neutral case. So, in this scenario, we have an inherent network whose nodes are themselves networks.\\

Now, we introduce the concept of basin of attraction to define the edges and weights of our inherent network. Note that for each solution $s$, there is a probability that $h(s) \in NN_{i}$. We denote $p_i(s)$ the probability $P(h(s) \in NN_{i})$.
We have for each solution $s \in S$, $\sum_{i=1}^{n} p_i(s) = 1$.

In non-neutral fitness landscapes where the size of each neutral network is $1$,
for each solution $s$, there exists only one neutral network (in fact one solution) $NN_i$ such that $p_i(s) = 1$. In this case, the basin of attraction of a local optimum neutral network $i$ is the set $b_i = \{ s \in S ~|~ p_i(s) = 1 \}$ which exactly correspond to our previous definition in~\cite{pre09}. We cannot use this definition in neutral fitness landscapes, but we can extend it in the following way:

\textbf{Basin of attraction:}
The basin of attraction of the local optimum neutral network $i$ is the set $b_i = \{ s \in S ~|~ p_i(s) > 0 \}$.
This definition is consistent with our previous definition \cite{gecco08,alife08} for the non-neutral case.\\

The size of each basin of attraction can now be defined as follows:

\textbf{Size of a basin of attraction:}
The size of the basin of attraction of a local optimum neutral network $i$ is $\sum_{s \in {\cal S}} p_i(s)$.\\

We are ready now to define the landscape's local optima network.

\textbf{Local optima network:}
The local optima network $G=(N,E)$ is the graph where the nodes are the local optima $NN$
and there is an edge between nodes $NN_i$ and $NN_j$ when
there are two solutions $s_i \in b_i$ and $s_j \in b_j$ such that $s_i \in V(s_j)$.\\

\textbf{Edge weight:}

We first reproduce the definition of edge weights for the non-neutral landscape \cite{alife08}:

\noindent For each solutions $s$ and $s^{'}$, let  $p(s
\rightarrow s^{'} )$ denote the probability that $s^{'}$ is a neighbor
of $s$, \textit{i.e.} $s^{'} \in V(s)$.
The probability that a configuration $s \in S$ has a neighbor in a basin $b_j$, is therefore:
$$
p(s \rightarrow b_j ) = \sum_{s^{'} \in b_j} p(s \rightarrow s^{'} )
$$

\noindent The total probability of going from basin $b_i$ to basin $b_j$ is the average over all $s \in b_i$ of the transition probabilities to solutions $s^{'} \in b_j$:

$$p(b_i \rightarrow b_j) = \frac{1}{\sharp b_i} \sum_{s \in b_i} p(s \rightarrow b_j )$$

Figure \ref{fig:nk62net} illustrates the complete network of a small non-neutral $NK$ landscape ($N=6$, $K=2$). The circles represent the local optima basins (with diameters indicating the size of basins),  and the weighted edges the transition probabilities as defined above.

For landscapes with neutrality, we have defined the probability $p_i(s)$ that a solution $s$ belongs to a basin $i$.
So, we can modify the previous definitions to consider neutral landscapes:
$$
p(s \rightarrow b_j ) = \sum_{s^{'} \in b_j} p(s \rightarrow s^{'} ) p_{j}(s^{'})
$$

\noindent and in the same way :

$$p(b_i \rightarrow b_j) = \frac{1}{\sharp b_i} \sum_{s \in b_i} p_i(s) p(s \rightarrow b_j )$$

\noindent where $\sharp b_i$ is the size of the basin $b_i$.

\noindent In the non-neutral case, we have $p_k(s)=1$ for all the configurations in the basin $b_k$. Therefore,  the definition of weights for the non-neutral case is consistent with the previous definition. Now, we are in a position to define the weighted local optima network:

\begin{figure} [!ht]
\begin{center}
\includegraphics[width=0.3\textwidth]{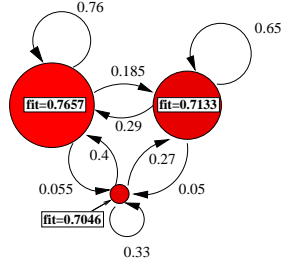} \\
\vspace{-0.3cm} \caption{Visualization of the weighted local optima network of a small $NK$ landscape ($N=6$, $K=2$). The nodes correspond to the local optima basins (with the diameter indicating the size of basins, and the label "fit", the fitness of the local optima). The  edges depict the transition probabilities between basins as defined in the text. \label{fig:nk62net}}
\end{center}
\end{figure}

\textbf{Weighted local optima network:}
The weighted local optima network $G_w=(N,E)$ is the graph where the nodes are the local optima neutral networks, and there is an edge $e_{ij} \in E$ with the weight $w_{ij} = p(b_i
\rightarrow b_j)$ between two nodes $i$ and $j$ if $p(b_i
\rightarrow b_j) > 0$.

According to our definition of edge weights, $w_{ij} = p(b_i \rightarrow b_j)$ may be different than $w_{ji} = p(b_j \rightarrow b_i)$. Thus, two weights are needed in general, and we have an oriented transition graph.

\section{Analysis of the local optima networks}
\label{analysis}


\subsection{Experimental setting}

In order to minimize the influence of the random creation of landscapes, we considered 30 different and independent landscapes for each parameter combinations: $N$, $K$ and $q$ or $p$. The measures reported, are the average of these 30 landscapes. We conducted our empirical study for $N$ = 18, which is the largest possible value of $N$ that allows the exhaustive extraction of $inherent$ networks. The remaining set of parameters explored are:   $K \in \{2, 4, 6, 8, 10, 12, 14, 16, 17 \}$, for $NK_q$ landscapes $q \in \{2,  4, 10 \}$, and for $NK_p$ landscapes, $p \in \{0.5, 0.8, 0.9 \}$.

\subsection{General Network Features}

This section describes some standard network features such as the number of nodes and edges, and the weight distribution of the edges. For all the combinations of landscape type and parameters, the measurements are the average of 30 independent landscape instances. When possible, we have also reported the data for the corresponding standard $NK$ landscape~\cite{alife08,pre09} in order to facilitate the comparison. In the figures, if not explicitly stated, the thick curves labeled $NK$ stand for the standard, non-neutral case.

\subsubsection{Number of nodes}

\begin{figure} [!ht]
\begin{center}
\includegraphics[width=0.4\textwidth]{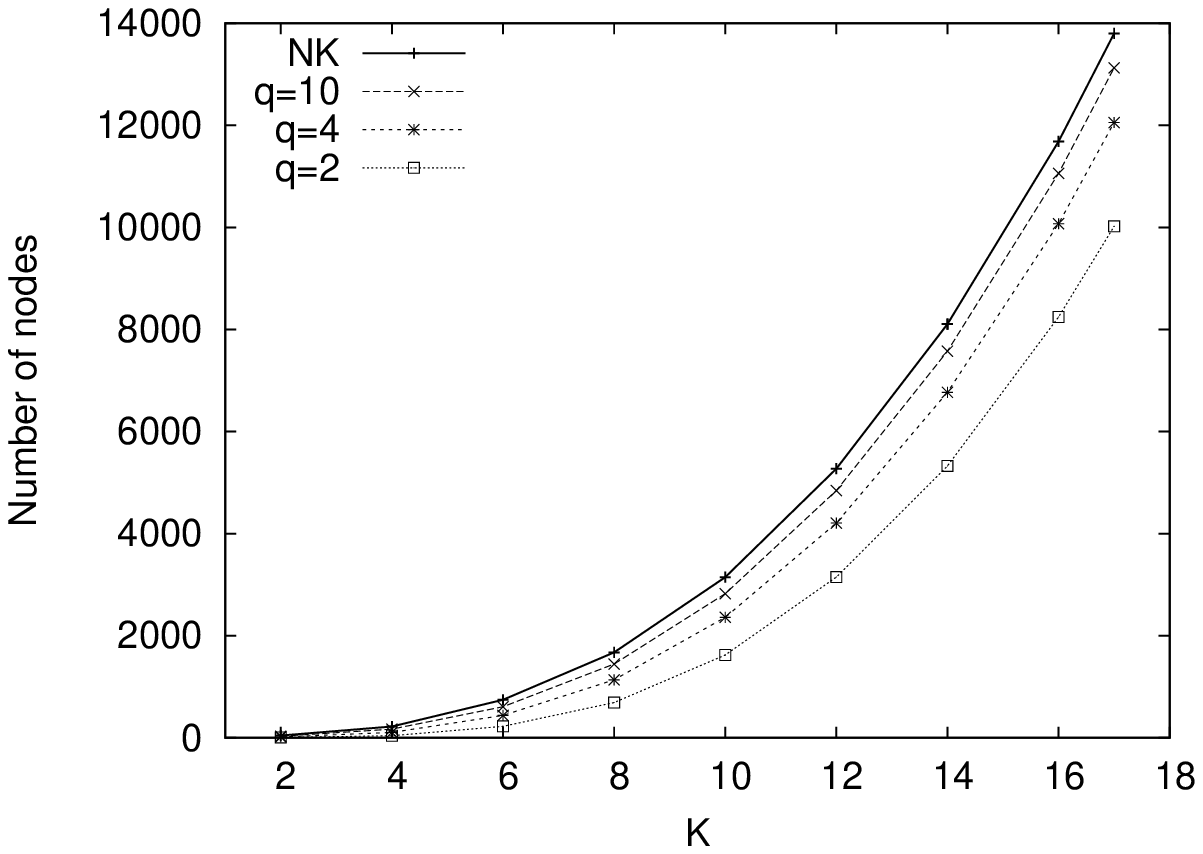} \\
\includegraphics[width=0.4\textwidth]{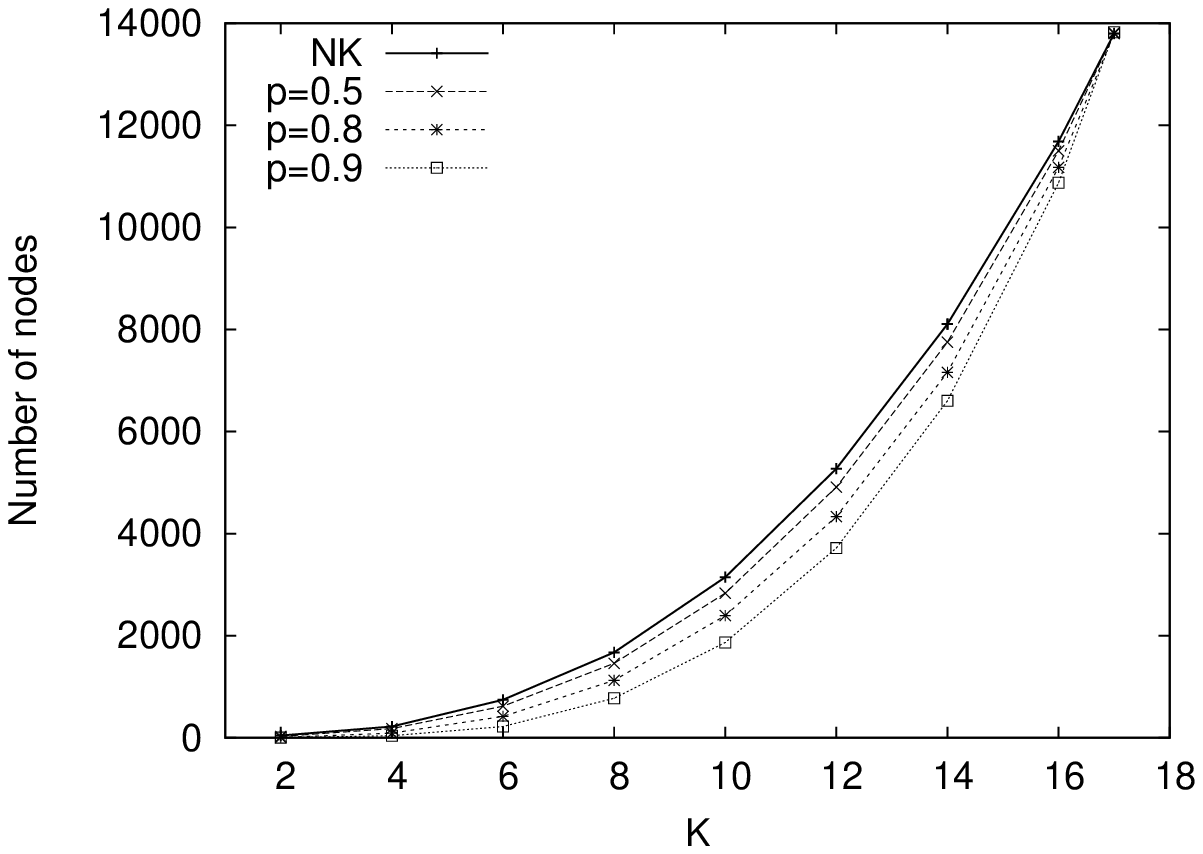} \\
\vspace{-0.3cm} \caption{Average number of nodes in the networks for
all the landscape parameters combinations. $NK_q$ landscapes (top), and $NK_p$ landscapes (bottom). Averages on 30 independent landscapes. Results for the standard $NK$ case are also shown
for comparison (thick lines).\label{fig:nodes}}
\end{center}
\end{figure}

Figure \ref{fig:nodes} shows the average of the number of nodes in the optima networks of both the $NK_q$ (top) and $NK_p$ (bottom) landscapes with all the combinations of parameters studied. Notice that the number of nodes increases rapidly as $K$ increases. Clearly, for given $N$ and $K$, the standard $NK$ landscape always has more nodes than the corresponding neutral version because the probability of changing fitness in non-neutral landscapes is higher than in neutral ones.
Therefore, for a given $K$, the number of nodes decreases with increasing neutrality.
All other things being equal, it is reasonable to assume that the search will be more difficult the larger the number of nodes. Therefore, as it is well known, the search is more difficult as $K$ increases, and for a given $K$, it will be more difficult when neutrality is low. In other words, an easier search will be expected for low $K$ and high neutrality.

\subsubsection{Number of edges}

\begin{figure} [!ht]
\begin{center}
\includegraphics[width=0.4\textwidth]{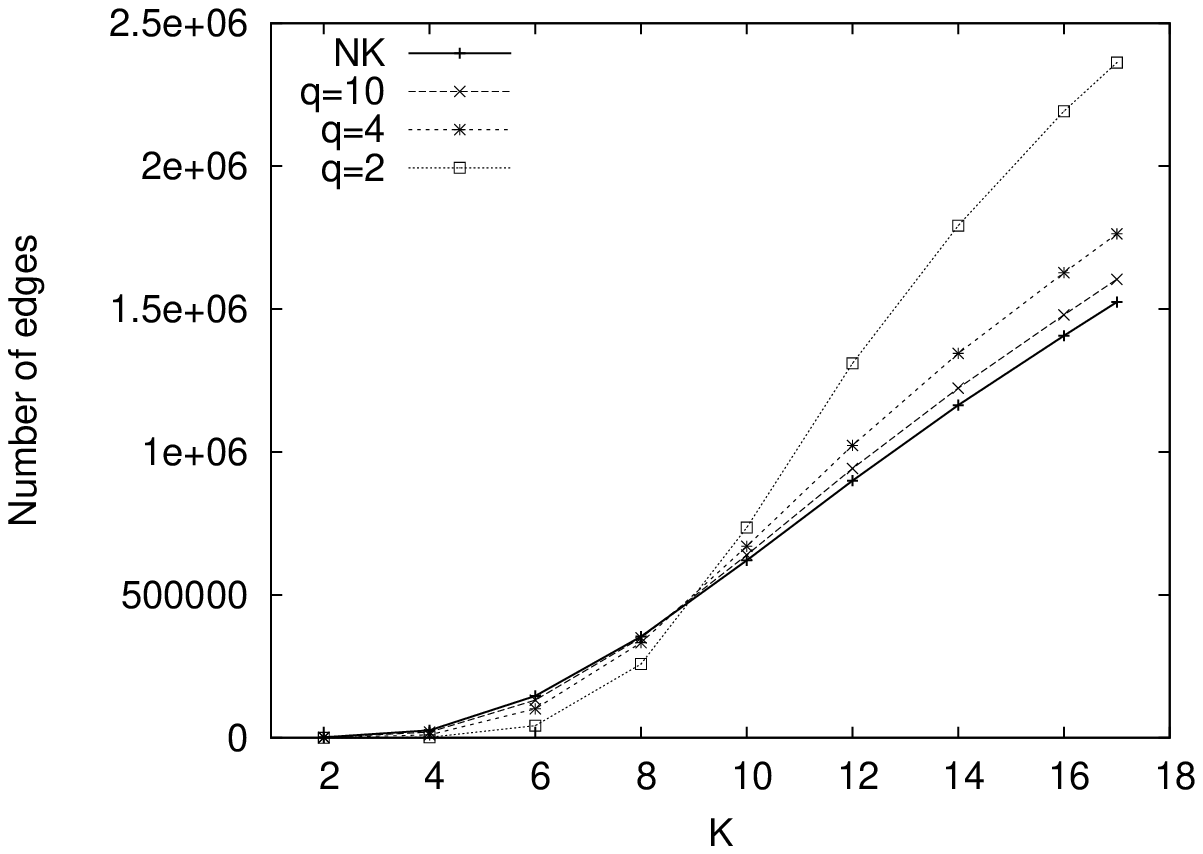} \\
\includegraphics[width=0.4\textwidth]{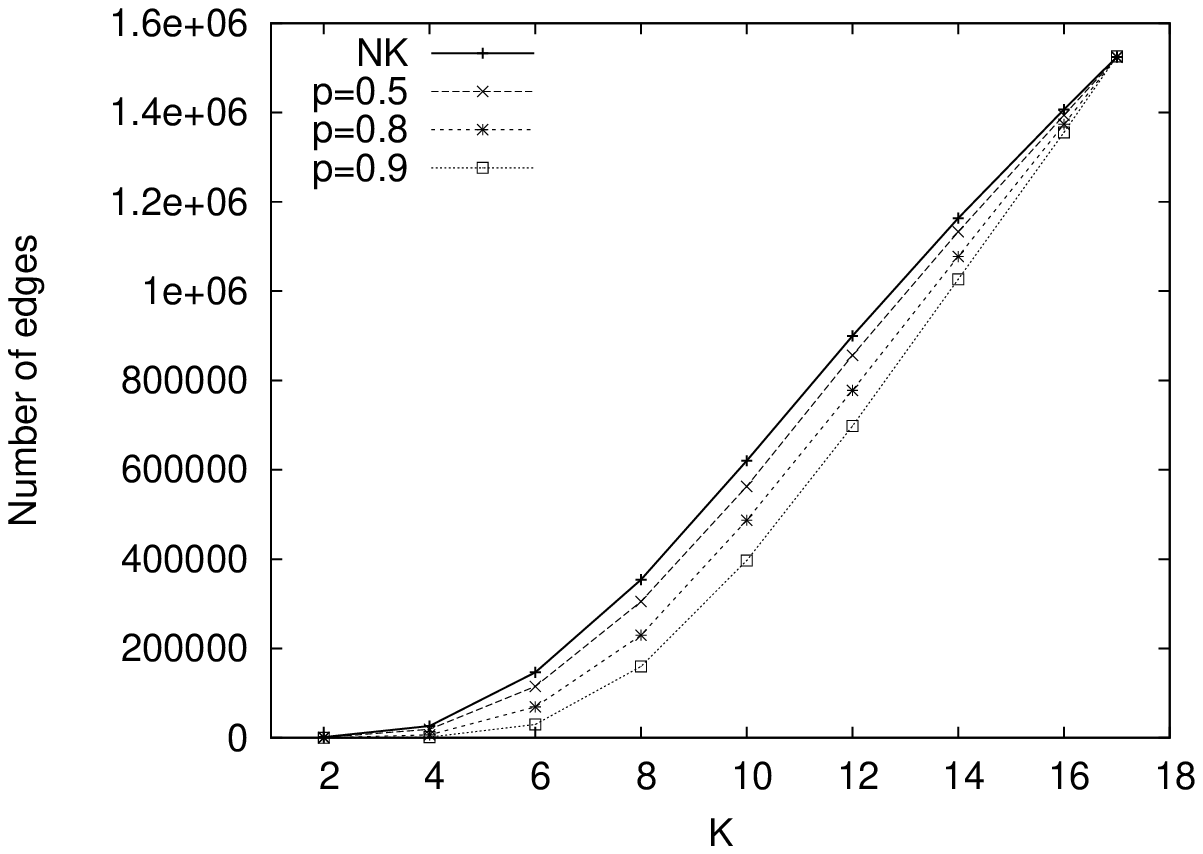} \\
\vspace{-0.3cm} \caption{Average number of edges with weight greater than 0,  for
all the landscape parameters combinations. $NK_q$ landscapes (top) and $NK_p$ landscapes (bottom).
Averages on 30 independent landscapes. The standard $NK$ data are also reported (thick lines). Note
the different scales on the y-axis.\label{fig:edges}}
\end{center}
\end{figure}

Similarly, Figure \ref{fig:edges} illustrates the average number of edges in the networks for both the $NK_q$ and $NK_p$ families of landscapes. Notice that the number of connections increases exponentially with increasing $K$. For the $NK_q$ landscape (Figure \ref{fig:edges}, bottom), the number of edges decreases with increasing neutrality for all $K$; whereas for $NK_p$ landscapes, this is true only for $K \leq 8$. In this case when $K > 8$ the trend is the opposite, that is the number of edges increases with increasing neutrality. The weight distribution results in the next subsection may help to clarify this finding.

\subsubsection{Weight Distribution}

For weighted networks, the weights are characterized by both the {\em weight distribution} $p(w)$ that any given edge has weight $w$, and the average of this distribution. In our study, for each node $i$, the total sum of weights from $i$ is equal to $1$. Therefore, an important measure is the weight $w_{ii}$ of self-connecting edges (i.e. configurations remaining in the same node).  We have the relation: $ w_{ii} + s_i = 1$.  $s_i$, the vertex {\em strength}, is defined as $s_i = \sum_{j \in V(i) \setminus \{i\}}  w_{ij}$ where the sum is over the set $V(i) \setminus  \{i\}$ of neighbors of $i$~\cite{bart05}. The strength of a node is a generalization of the node's connectivity giving information about the number and importance of the edges.

\begin{figure} [!ht]
\begin{center}
\includegraphics[width=0.4\textwidth]{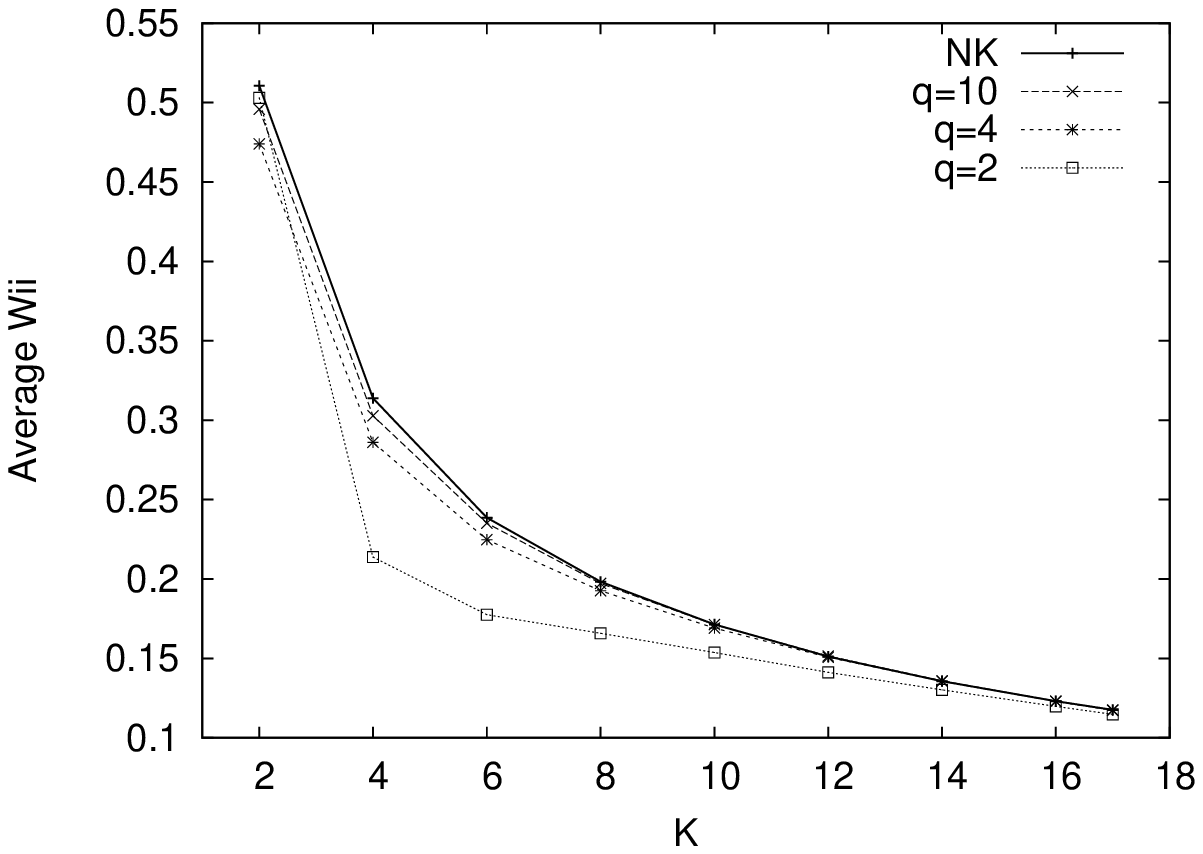} \\
\includegraphics[width=0.4\textwidth]{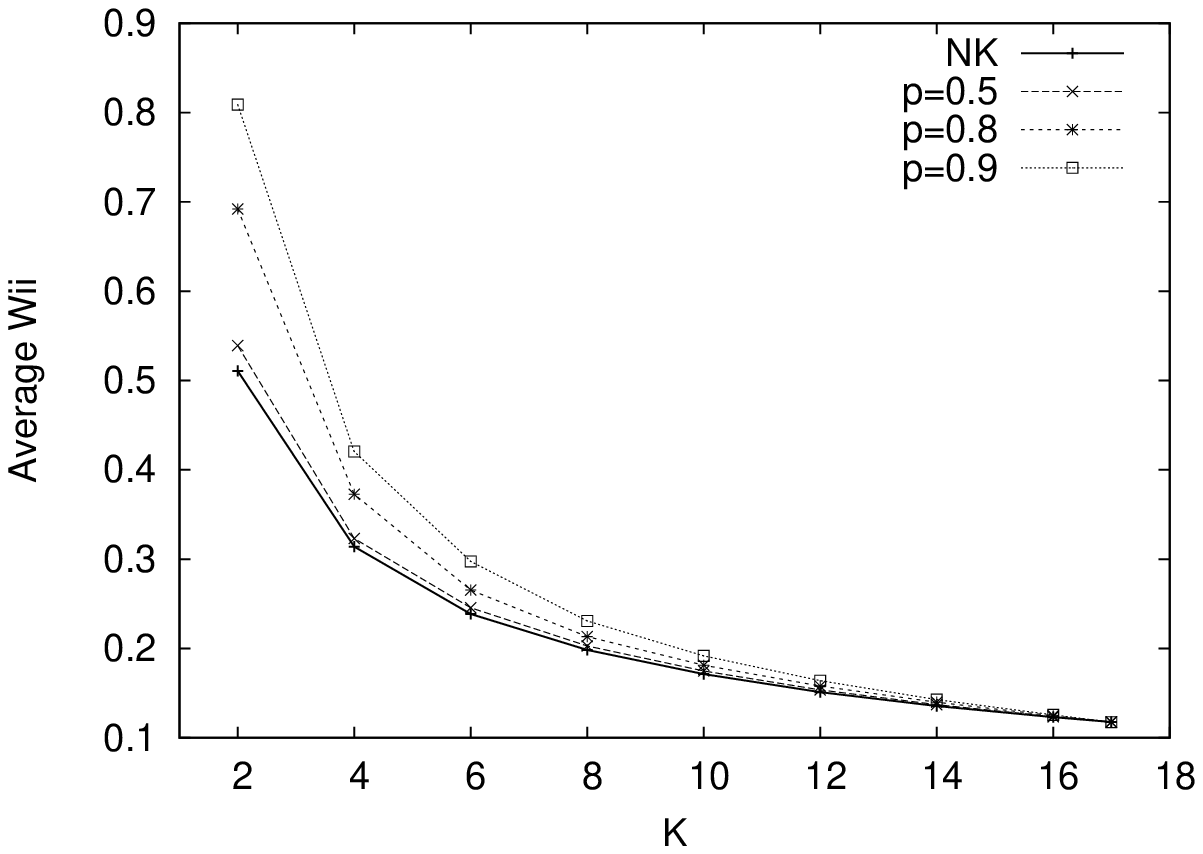} \\
\vspace{-0.3cm} \caption{Average weight $w_{ii}$ according to the
parameters for $NK_q$ landscapes (top) and $NK_p$ landscapes (bottom).
Averages on 30 independent landscapes. \label{fig:wii}}
\end{center}
\end{figure}

\begin{figure}[!ht]
\begin{center}
\begin{tabular}{cc}
\includegraphics[width=0.4\textwidth]{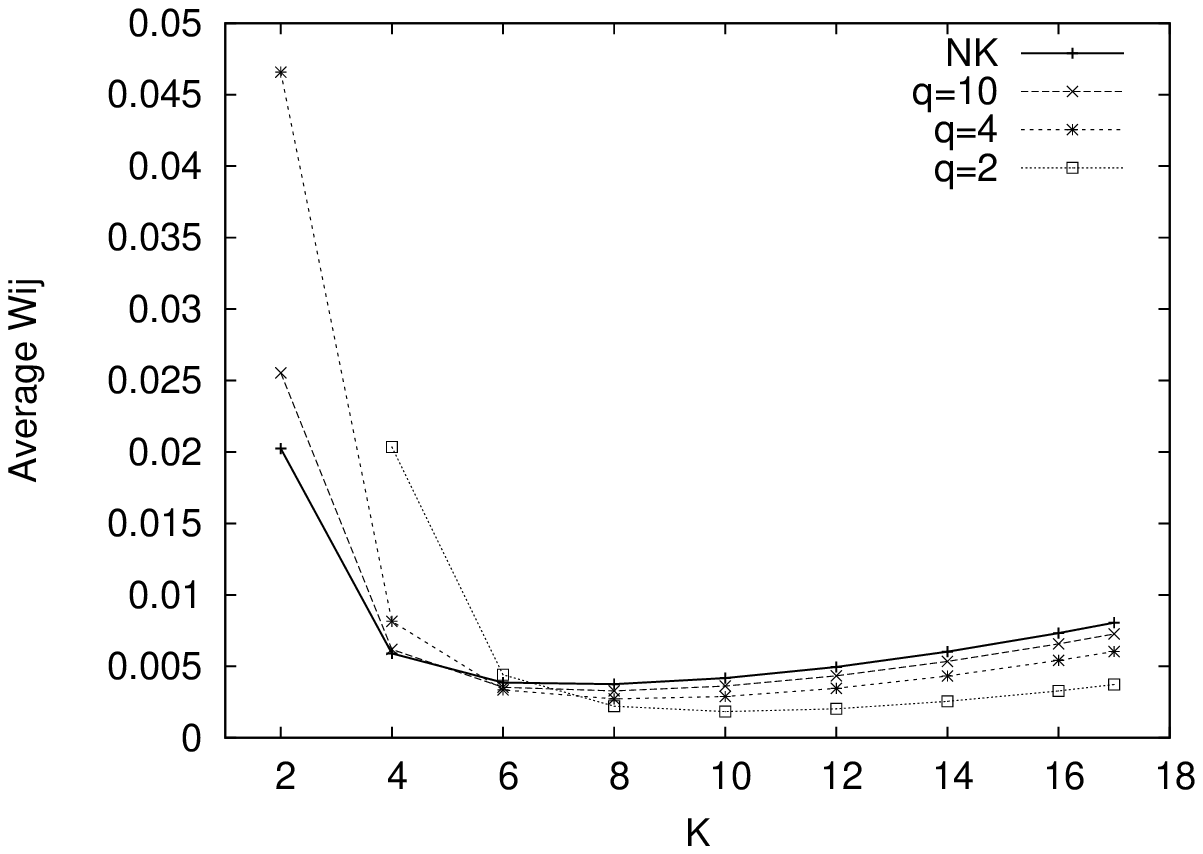} \\
\includegraphics[width=0.4\textwidth]{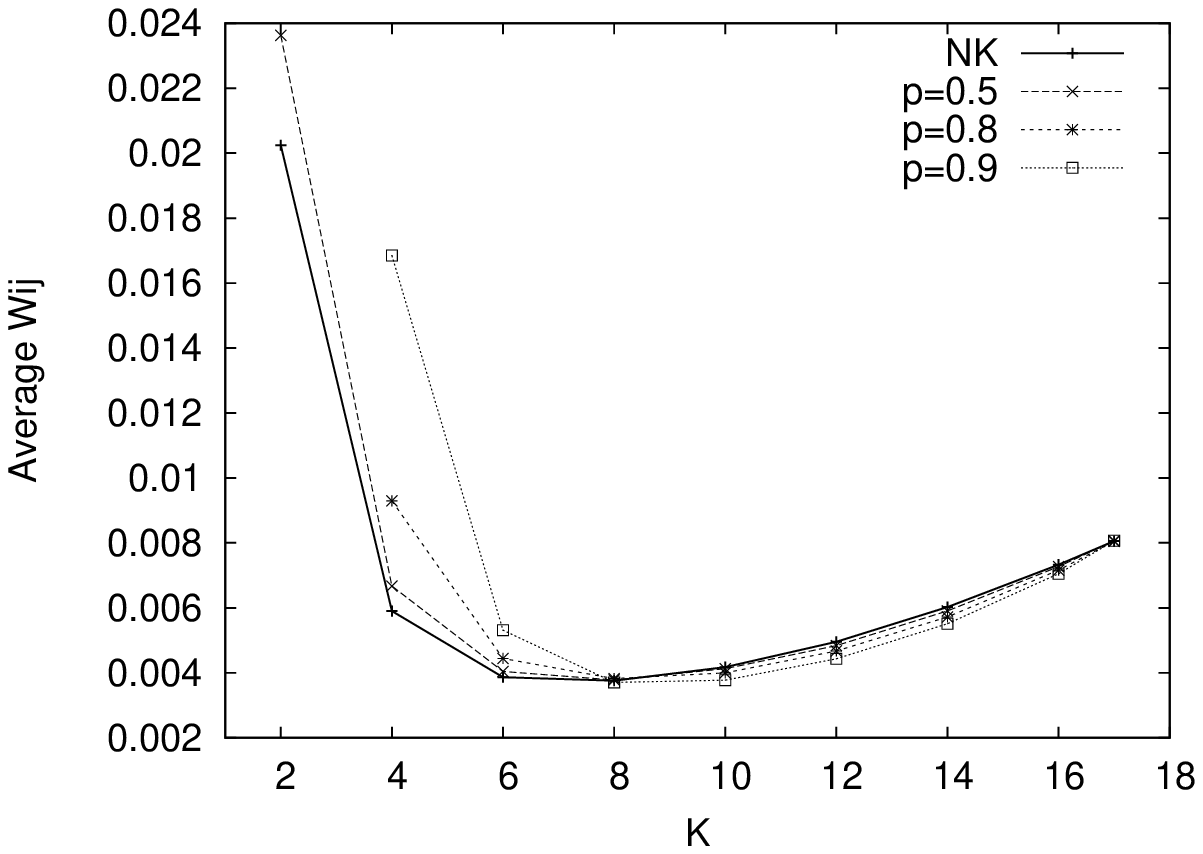} \\
\end{tabular}
\vspace{-0.3cm}
\caption{Average of the outgoing weights $w_{ij}$ where $i \not= j$, for $NK_q$ landscapes (top) and $NK_p$ landscapes (bottom). Averages on 30 independent landscapes. \label{fig:Wij}}
\end{center}
\end{figure}

Figure \ref{fig:wii} shows the averages, over all the nodes in the network, of the weights $w_{ii}$ (i.e. the probabilities of remaining in the same basin after a hill-climbing from a mutation of one configuration in the basin). On the other hand, Figure~\ref{fig:Wij} shows the empirical average of weights $w_{ij}$ with $i \not= j$. It is clear from these results that jumping into another basin is much less likely than walking around in the same basin (approximately by an order of magnitude). Notice that for both types of neutral landscapes, the weights to remain in the same basin, $w_{ii}$ (fig.~\ref{fig:wii}), decrease with increasing $K$, which is also the trend followed in standard $NK$ landscapes. The weights to get to another basins (fig.~\ref{fig:Wij}) also decrease with increasing $K$ up to  $K=8$, thereafter they seem to remain constant or increase slightly. This can be explained as follows, as the number of basins increases non-linearly with increasing $K$, the probability to get to one particular basin decreases.

The trend with regards to neutrality is more complex, and it is different for the two families of neutral landscapes. On the $NK_q$ landscape, for a fixed $K$, the average weight to stay in the same basin decreases with increasing neutrality (fig.~\ref{fig:wii}, top); whereas the opposite happens on the $NK_p$ landscape, that is, the average weight to stay in the same basin increases with neutrality (fig.~\ref{fig:wii}, bottom). The trend of the weights to get to another basin (fig.~\ref{fig:Wij}) is similar for both families of landscapes. It changes when $K=8$: for $K < 8$ it increase with neutrality, while for $K>8$ it is nearly constant. Therefore, neutrality increases the probability that a given configuration escapes its basin and gets to another basin; but neutrality also increases the number of basins to which the current configuration is linked.

\begin{figure}[!ht]
\begin{center}
\begin{tabular}{cc}
\includegraphics[width=0.4\textwidth]{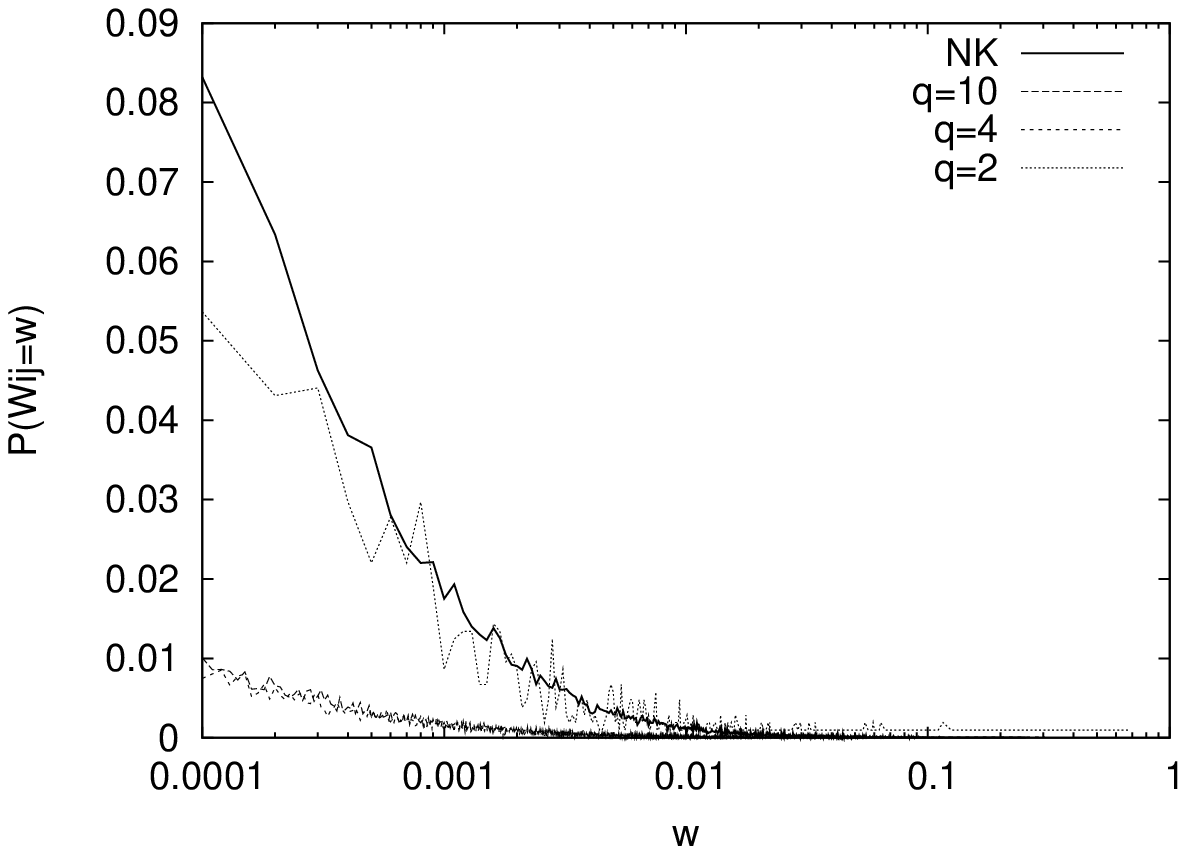} \\
\includegraphics[width=0.4\textwidth]{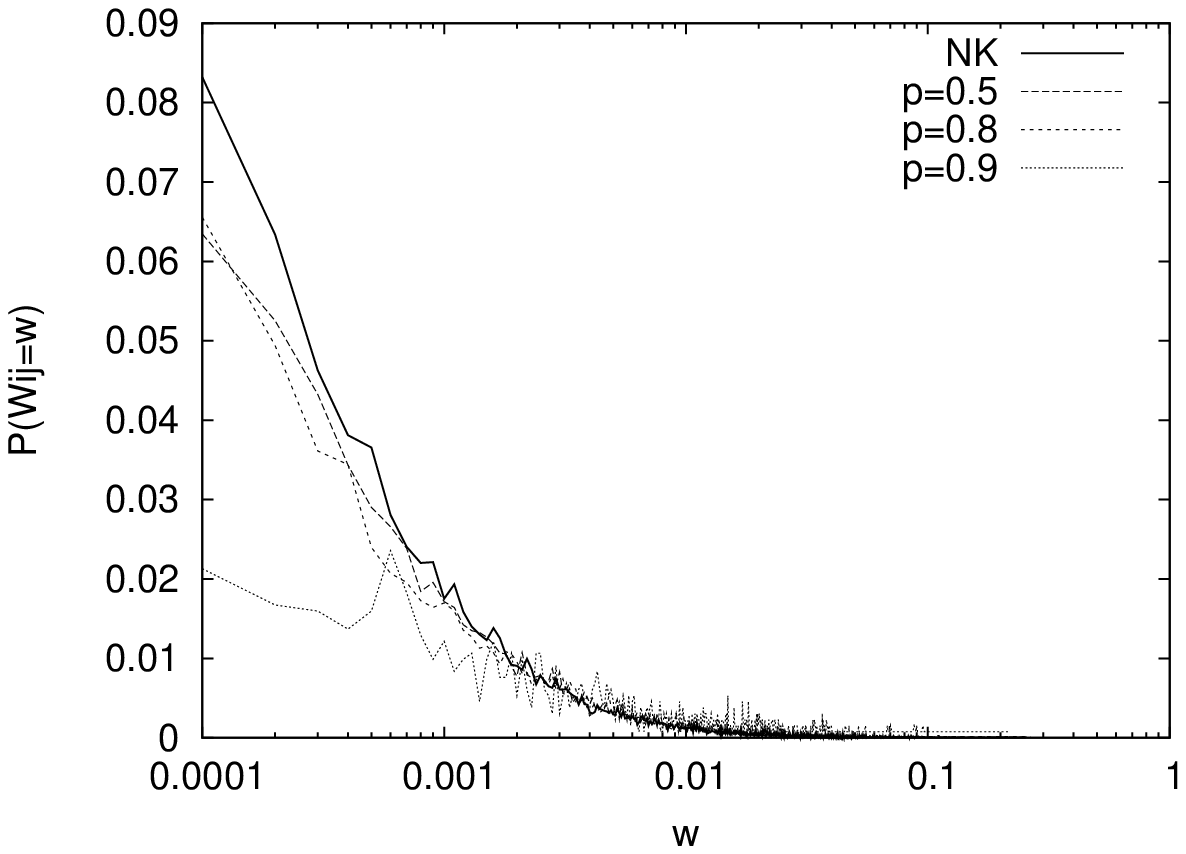} \\
\end{tabular}
\vspace{-0.3cm}
\caption{Probability distribution of the network
weights $w_{ij}$ for outgoing edges with $j\not=i$ in logscale on x-axis, for $NK_q$ landscapes (top) and $NK_p$ landscapes (bottom). Averages on 30 independent landscapes.\label{fig:Wij_distri}}
\end{center}
\end{figure}

The general network features discussed in this section are related to the search difficulty on the corresponding landscapes\footnote{The Appendix reports an empirical study exploring the effect of neutrality on the search difficulty for a standard evolutionary algorithm.}, since they reflect both the number of basins, and the ability to navigate the landscapes.

\subsection{Basins of attraction}

Besides the local optima networks, it is useful to describe the associated basins of attraction as they play a key role in heuristic search algorithms. Furthermore, some characteristics of the basins can be related to the local optima network features. The notion of the basin of attraction of a local maximum has been presented in section \ref{defs}. We have exhaustively computed the size and number of the basins of all the neutral landscapes under study.

\subsubsection{Number of basins of a given size}

\begin{figure*} [!ht]
\begin{center}
\begin{tabular}{cc}
\includegraphics[width=0.4\textwidth]{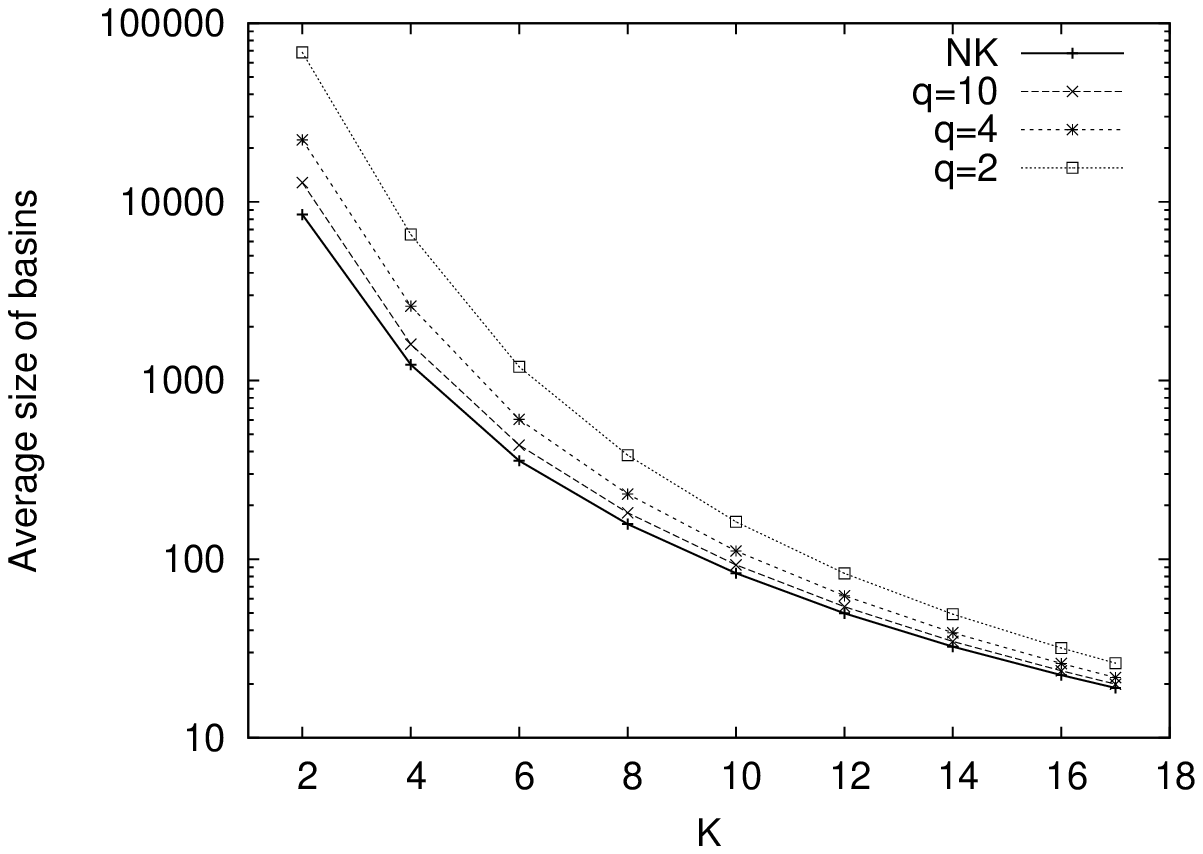} &
\includegraphics[width=0.4\textwidth]{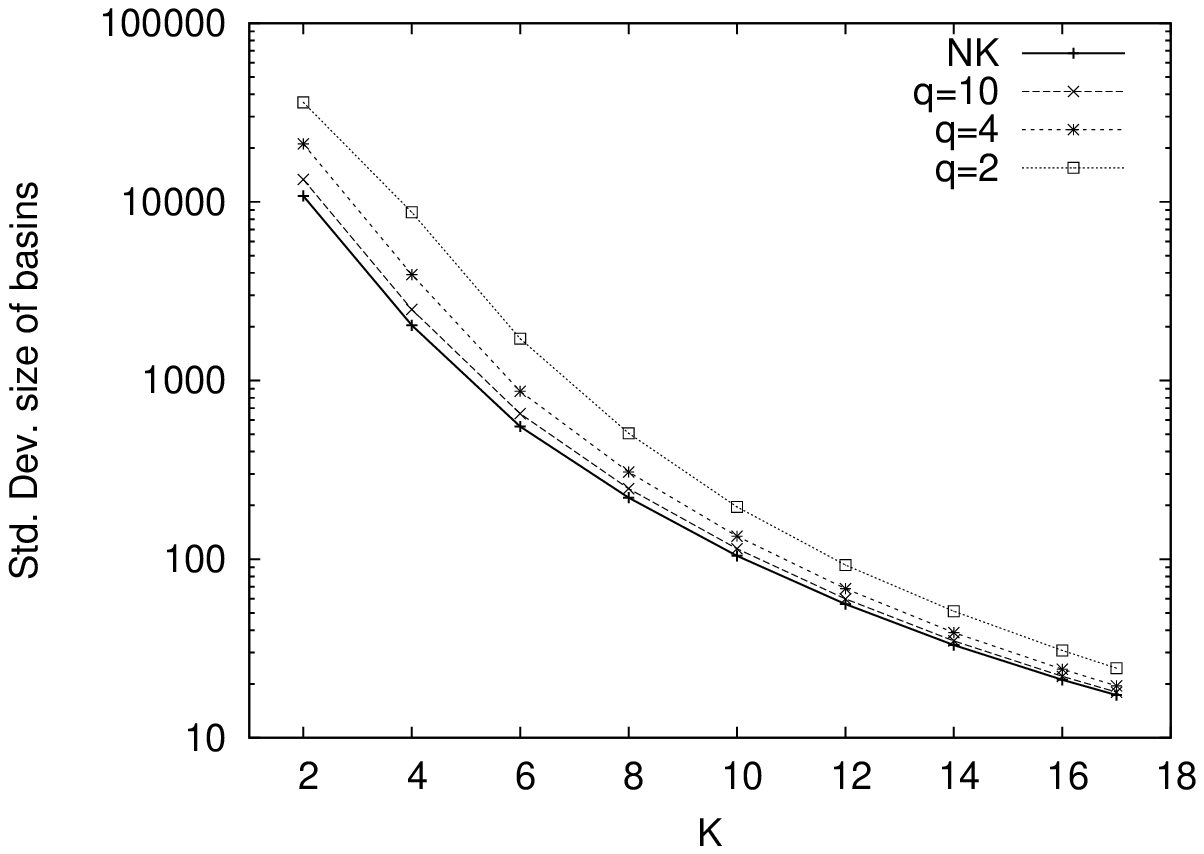} \\
\includegraphics[width=0.4\textwidth]{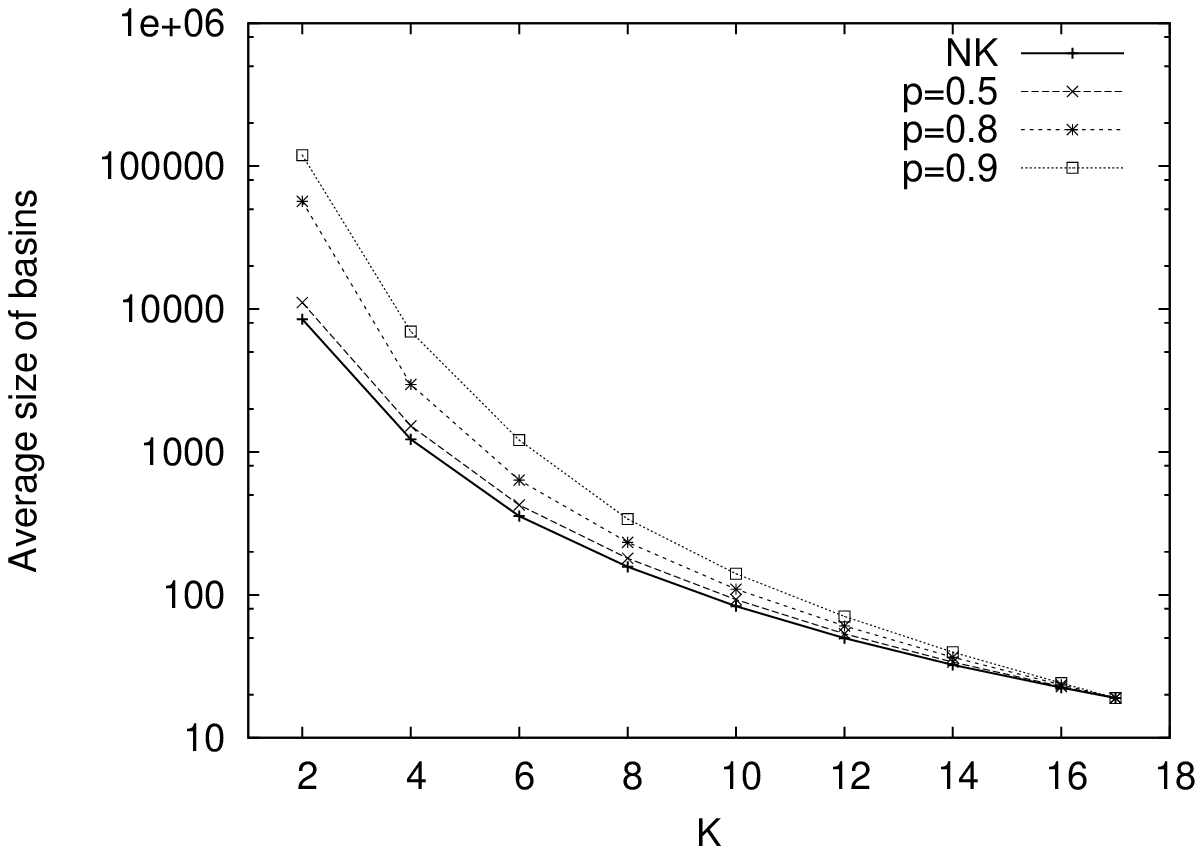} &
\includegraphics[width=0.4\textwidth]{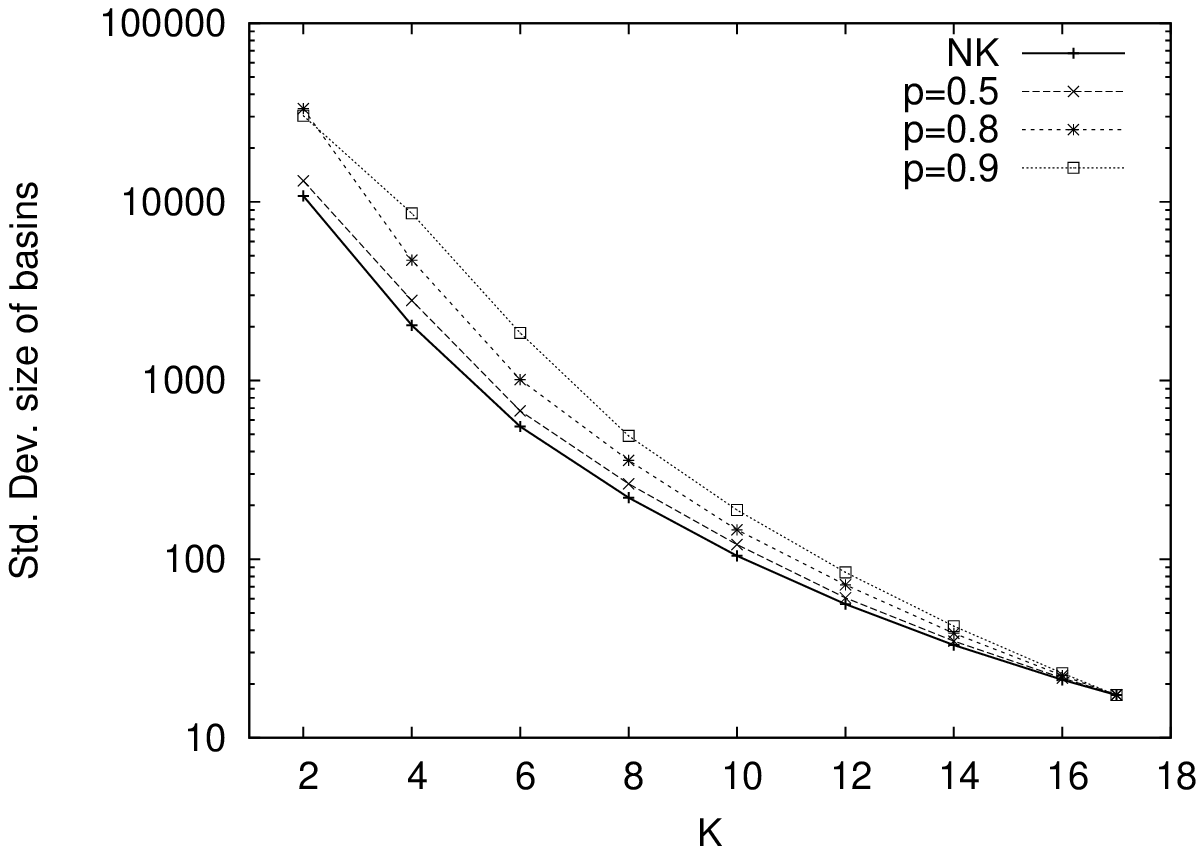} \\
\end{tabular}
\end{center}
\caption{Average (left) and standard deviation (right) of distribution of sizes for $NK_q$ landscapes $K=4$ (top) and for $NK_p$ landscapes $K=4$ (bottom). Averages on 30 independent landscapes.}\label{fig:distri-size}
\end{figure*}

Fig.~\ref{fig:distri-size} shows the average size (left) and standard deviation (right) of the basins for all the studied landscapes (averaged over the 30 independent instances in each case).
Notice that size of basins decreases exponentially with increasing $K$. They also decrease when  neutrality decreases, being smallest for non-neutral $NK$ landscapes, as one would expect intuitively. The standard deviations show the same behaviour as the average. It decreases exponentially with increasing $K$ and also decreases when the neutrality decreases.

\begin{figure} [!ht]
\begin{center}
\begin{tabular}{c}
\includegraphics[width=0.4\textwidth]{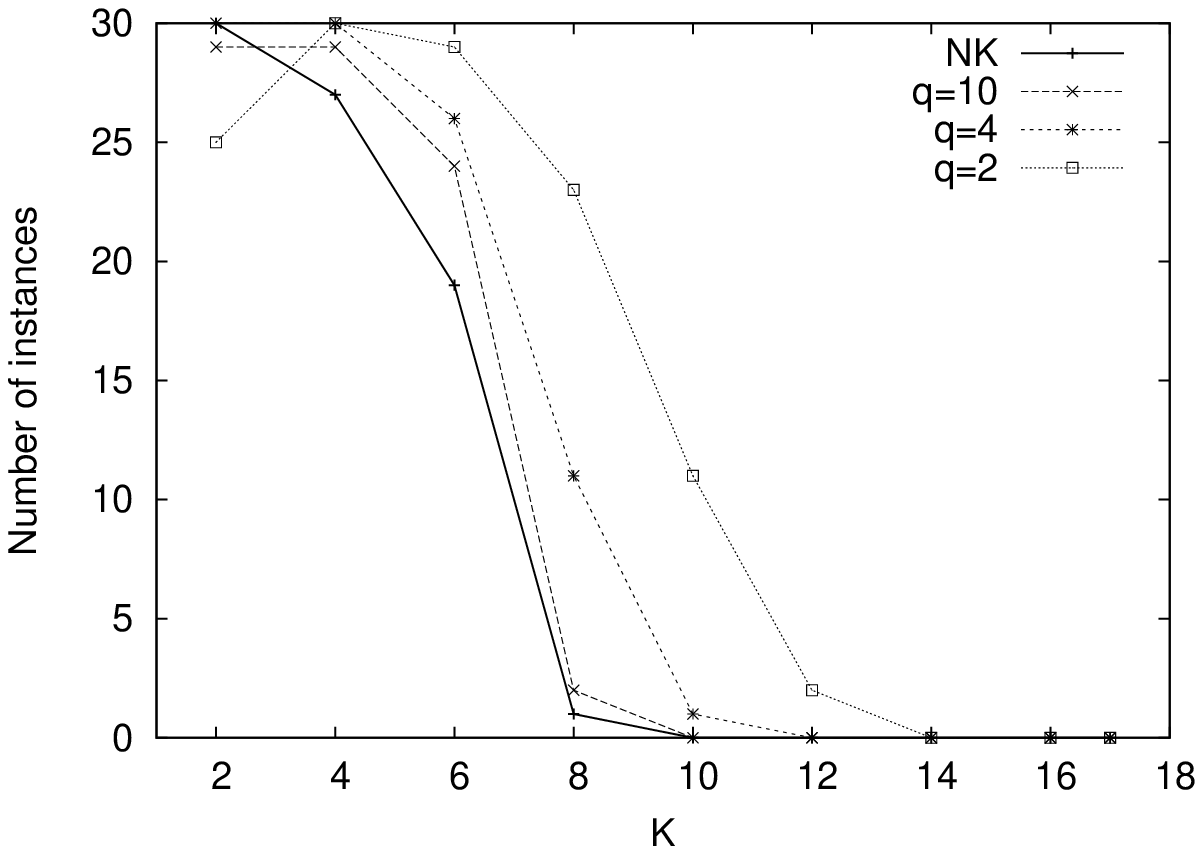} \\
\includegraphics[width=0.4\textwidth]{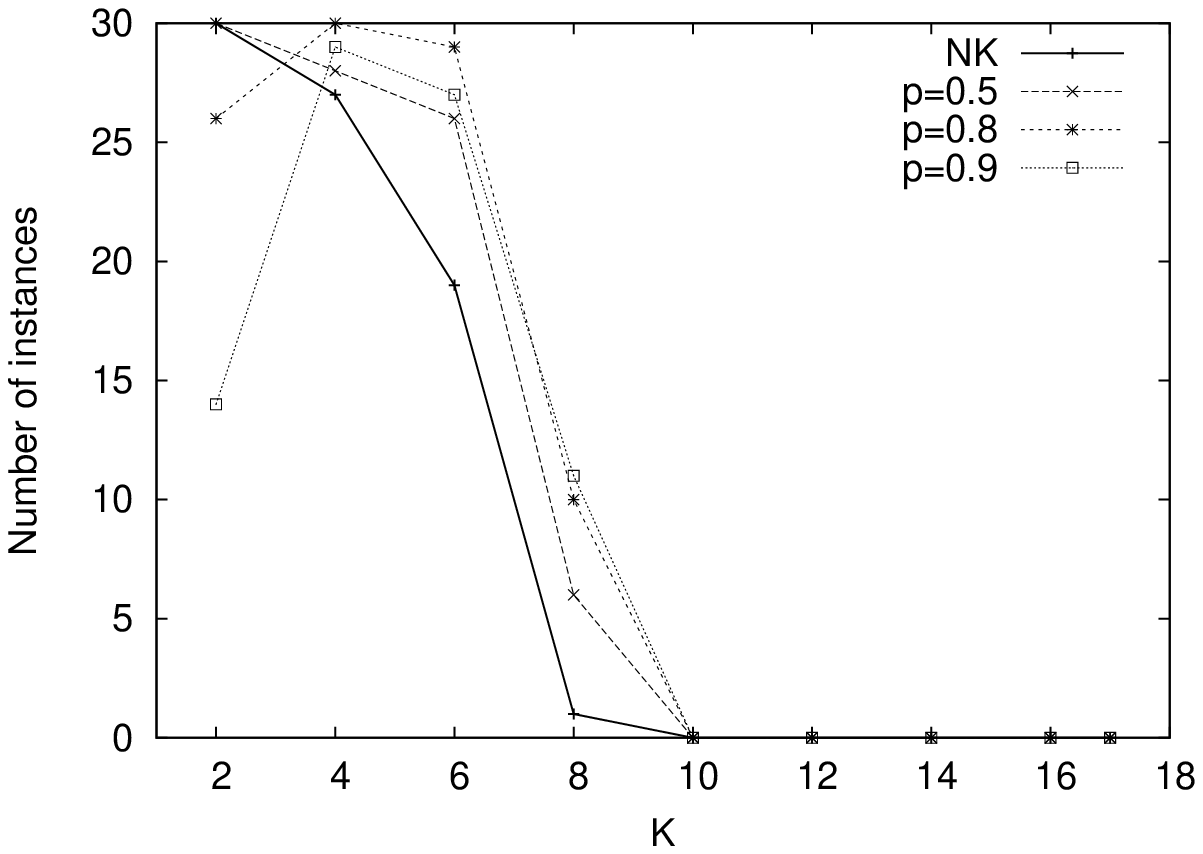} \\
\end{tabular}
\end{center}
\caption{Number of landscape instances (over the 30 independent instances) where the size distribution is a log-normal distribution according to the Shapiro-Wilk normality test at level of $1 \%$ for $NK_q$ landscapes (top) and for $NK_p$ landscapes (bottom).}\label{fig:distri-size-fit}
\end{figure}

Using the Shapiro-Wilk normality test~\cite{shapiroWilk} we confirmed that some distributions of basin's sizes can be fitted by a log-normal law when $K$ is low. Fig~\ref{fig:distri-size-fit} shows the number of landscape instances where the size distribution can be fitted by a log-normal distribution according to the statistical test at level of $1 \%$. The number $30$ on the y-axis means that for all the instances studied the size distribution can be fitted by a log-normal. For the non-neutral $NK$ landscapes when $K \leq 6$, nearly all the size distribution are log-normal.

For $K \geq 4$, the neutrality increases the number of log-normal distributions.
Again the influence of neutrality on the two types of landscapes is not the same:
for $NK_q$ landscapes, the number of log-normal distributions increases when there is more neutrality whereas, the number of log-distribution is not maximal for the more neutral $NK_p$ landscapes. For large $K$, the average size of basins is very small (Fig.~\ref{fig:distri-size} left). In this case, the size distributions are not log-normal, and become very narrow.
Few different sizes exist and those are very small. This confirms the ruggedness of the landscape when $K$ is very large even when there is some neutrality. The log-normal distribution implies that the majority of basins have a size close to average; and that there are few basins with larger than average size. We will see that this may be related to the search difficulty on the underlying landscape.

\subsubsection{Fitness of local optima}

\begin{figure*} [!ht]
\begin{center}
\begin{tabular}{cc}
\includegraphics[width=0.4\textwidth]{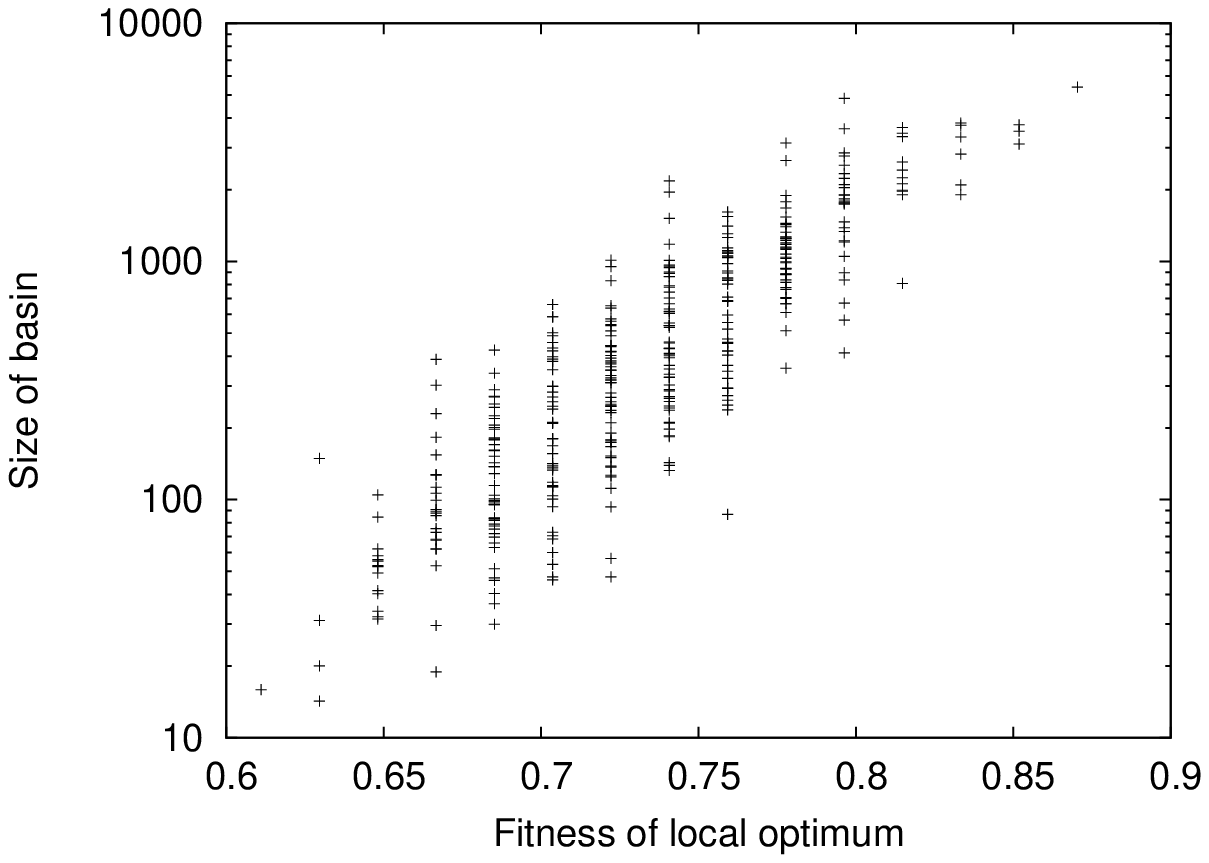} &
\includegraphics[width=0.4\textwidth]{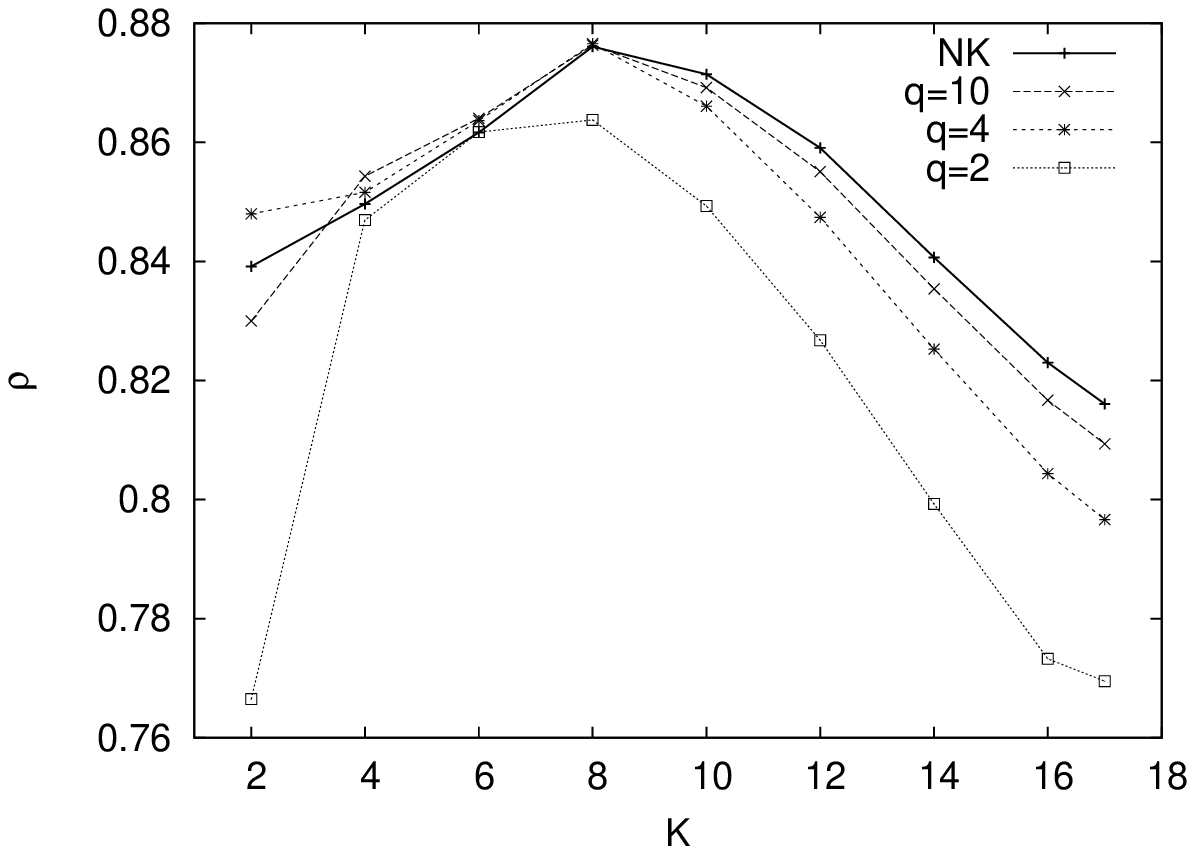} \\
\includegraphics[width=0.4\textwidth]{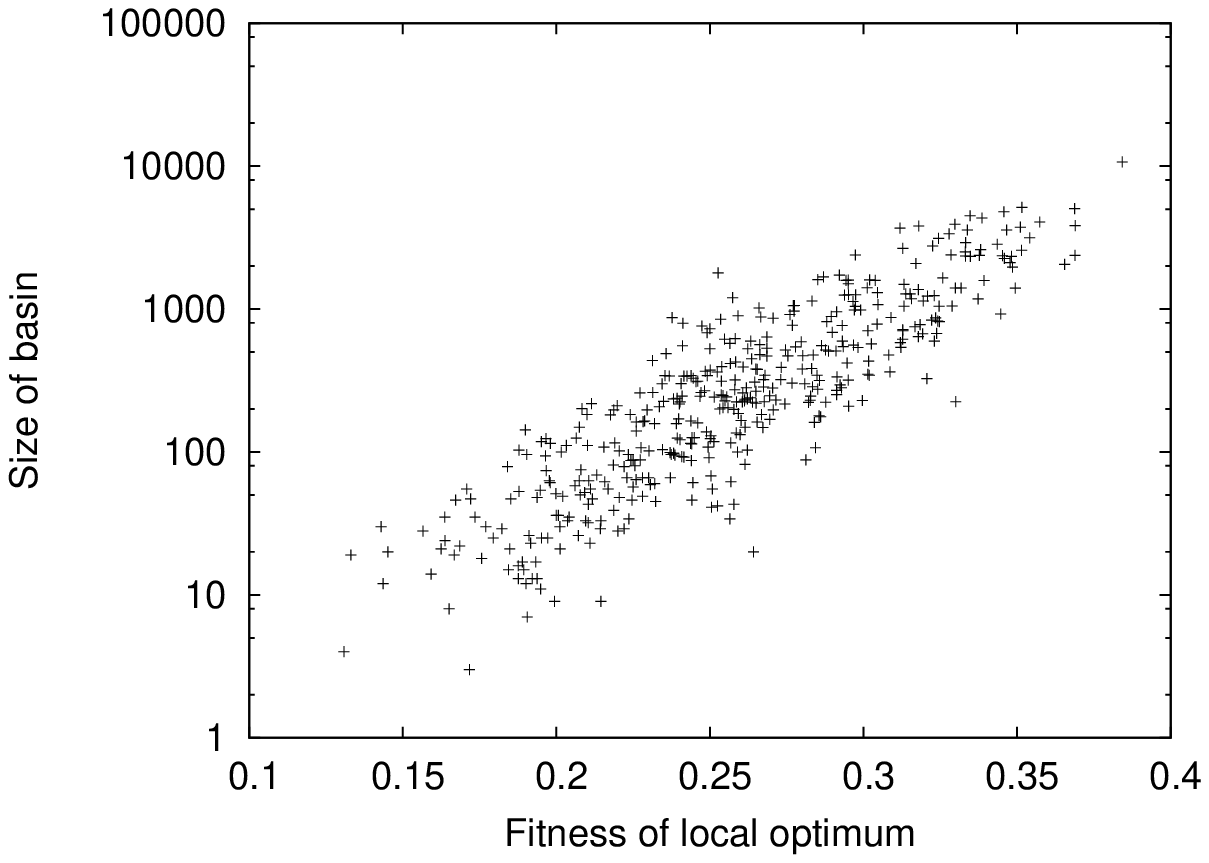} &
\includegraphics[width=0.4\textwidth]{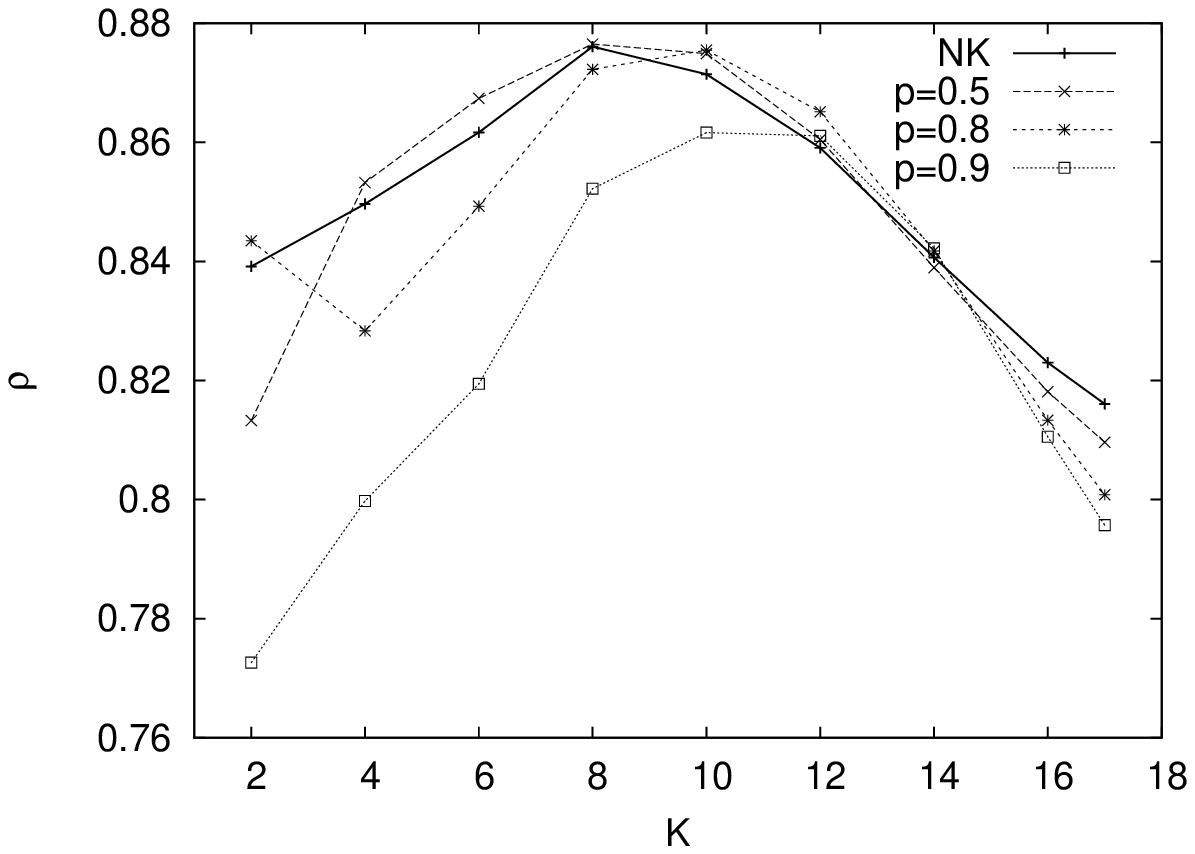} \\
\end{tabular}
\end{center}
\caption{Correlation between the fitness of local optima and their corresponding basin sizes (in log) for $NK_q$ landscapes (top) and $NK_p$landscapes (bottom). Two representative instances with $K=6$, $q=4$ and $p=0.8$ (left) and the average of correlation coefficient on $30$ independent landscapes for each parameters (right).
Averages on 30 independent landscapes.\label{fig:cor_fit-size}}
\end{figure*}

The scatter-plots in Fig.~\ref{fig:cor_fit-size} (left) illustrate the correlation between the basin sizes (in logarithmic scale) and their fitness values, for two representative landscape instances (with $K=6$, $q=3$ and $p=0.8$ ). Fig.~\ref{fig:cor_fit-size} (right) reports the correlation coefficients $\rho$ for all combinations of landscape types and its parameters. Notice that the correlations are positive and high, which implies that the larger basins have the higher fitness value. Therefore, the most interesting basins are also the larger ones! This may be surprising, but consider that our results on basin sizes show that the size differences between large and small decreases with increasing epistases. In consequence, with increasing ruggedness the difficulty to find the basin with higher fitness, also increases. Notice also that the correlations increase with $K$, up to $K = 8$ and then they decrease. Fig.~\ref{fig:cor_fit-size}, also illustrates that neutrality decreases the correlation between basin sizes and their fitness values. In other words, the size of basins is less related to the fitness of their local optima when neutrality is present. But, as we have discussed before, basins are larger in size and smaller in number with increasing neutrality.

\subsubsection{Global optimum basin size}

\begin{figure} [!ht]
\begin{center}
\begin{tabular}{c}
\includegraphics[width=0.4\textwidth]{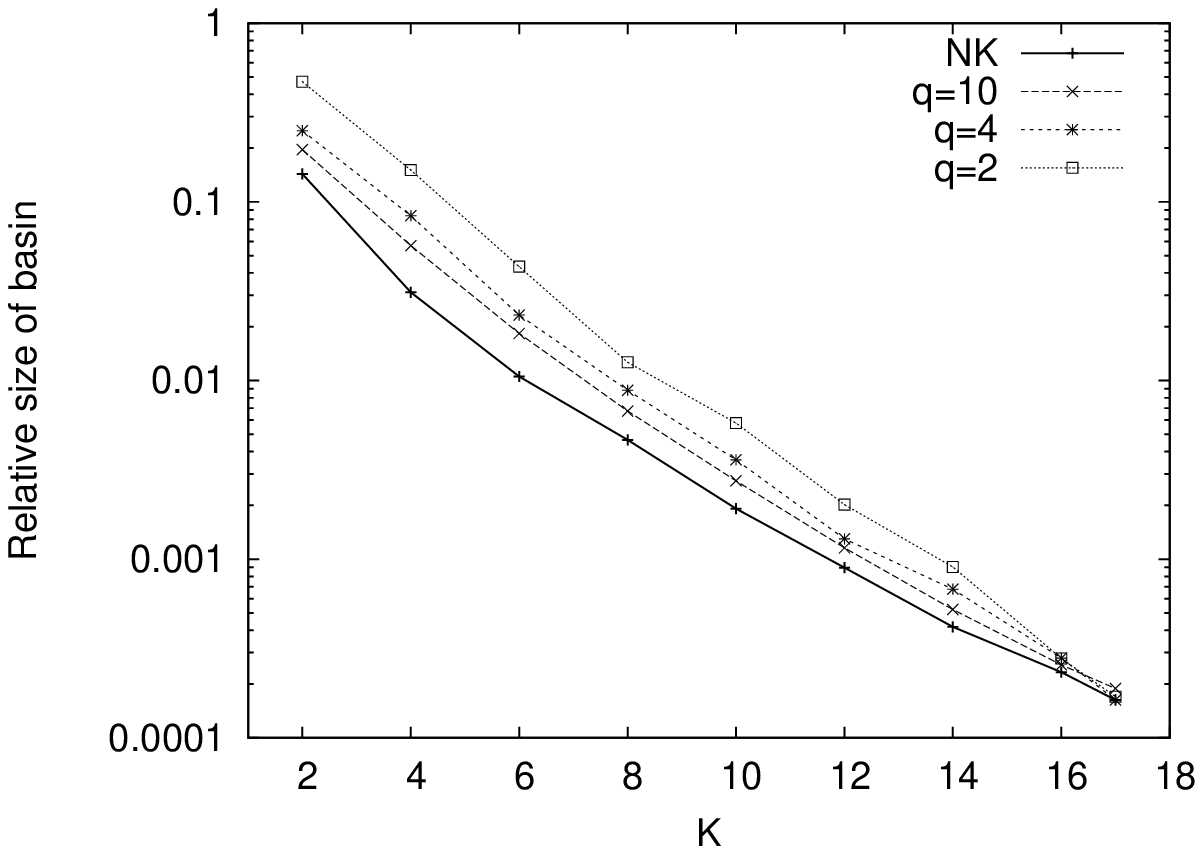} \\
\includegraphics[width=0.4\textwidth]{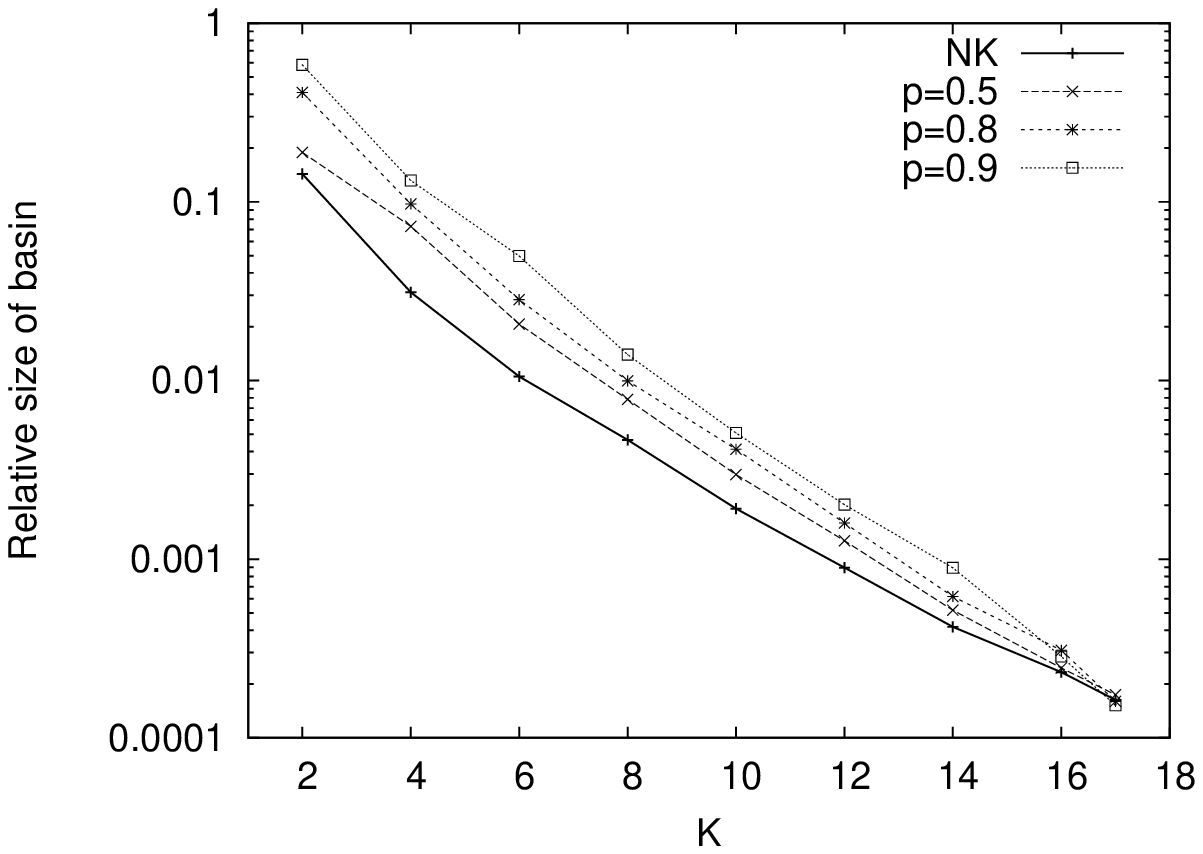} \\
\end{tabular}
\caption{Average of the relative size of the basin corresponding to
the global maximum for each K and neutral parameter
over 30 independent landscapes (top $NK_q$ and bottom $NK_p$).\label{nbasins}}
\end{center}
\end{figure}

In Fig.~\ref{nbasins} we plot the average size of the basin corresponding to the global maximum for all combinations of landscape types and its parameters. The results clearly show that the size exponentially decreases when $K$ increases. This agrees with our previous  results on standard $NK$ landscapes \cite{gecco08,alife08}. With respect to neutrality the size of the global maximum basin increases with increasing neutrality.

\subsection{Advanced network features}
\label{advancedFeatures}

In this section, we study the weighted clustering coefficient, the average path length between nodes, and the disparity of the local optima networks.

\subsubsection{Clustering Coefficient}

The standard  clustering coefficient~\cite{newman03} does not consider weighted edges. We thus use the {\em weighted clustering} measure proposed by~\cite{bart05}, which combines the topological
information with the weight distribution of the network:

$$c^{w}(i) = \frac{1}{s_i(k_i - 1)} \sum_{j,h} \frac{w_{ij} + w_{ih}}{2} a_{ij} a_{jh} a_{hi}$$
where $s_i = \sum_{j \not= i} w_{ij}$, $a_{nm} = 1$ if $w_{nm} > 0$,
$a_{nm} = 0$ if $w_{nm} = 0$ and $k_i = \sum_{j \not= i} a_{ij}$.

For each triple formed in the neighborhood of the vertex $i$, $c^{w}(i)$ counts the weight of the two participating edges of the vertex $i$. $C^w$ is defined as the weighted clustering coefficient
averaged over all vertices of the network.

\begin{figure} [!ht]
\begin{center}
\begin{tabular}{cc}
\includegraphics[width=0.4\textwidth]{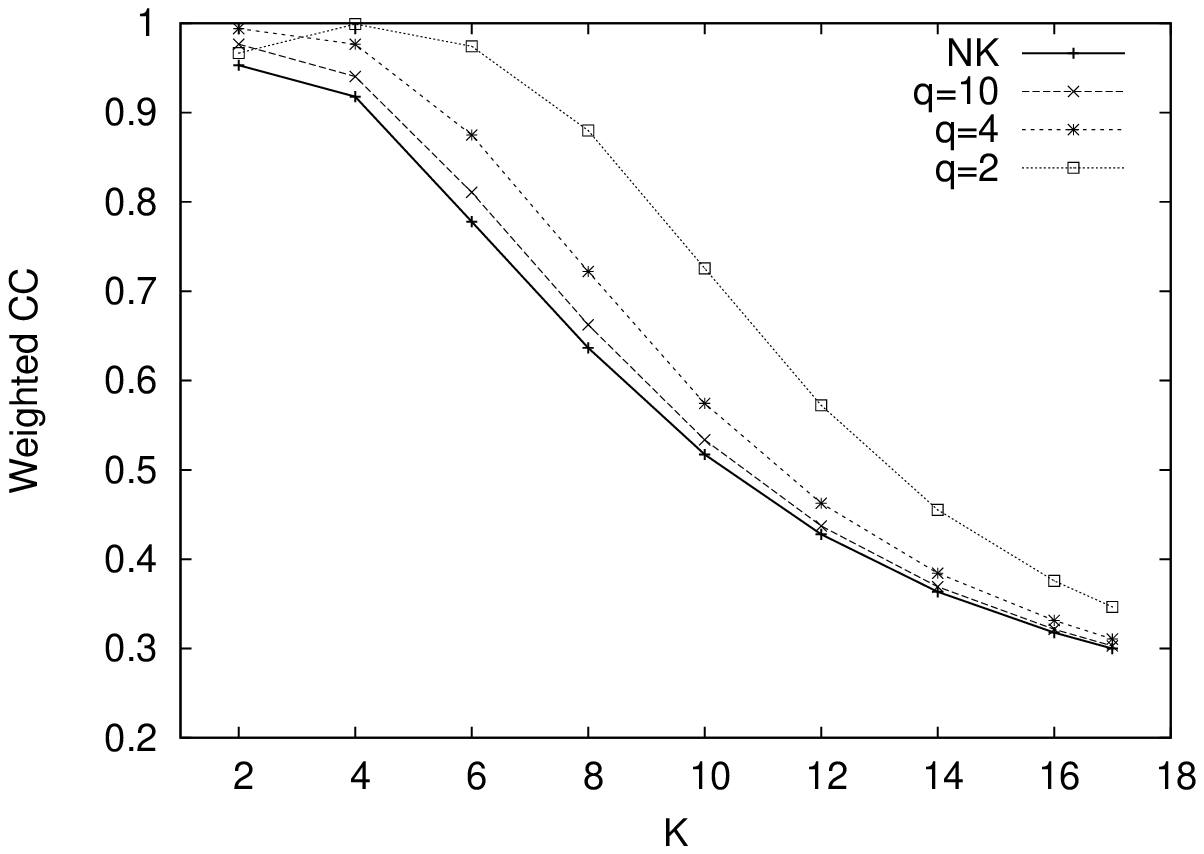} \\
\includegraphics[width=0.4\textwidth]{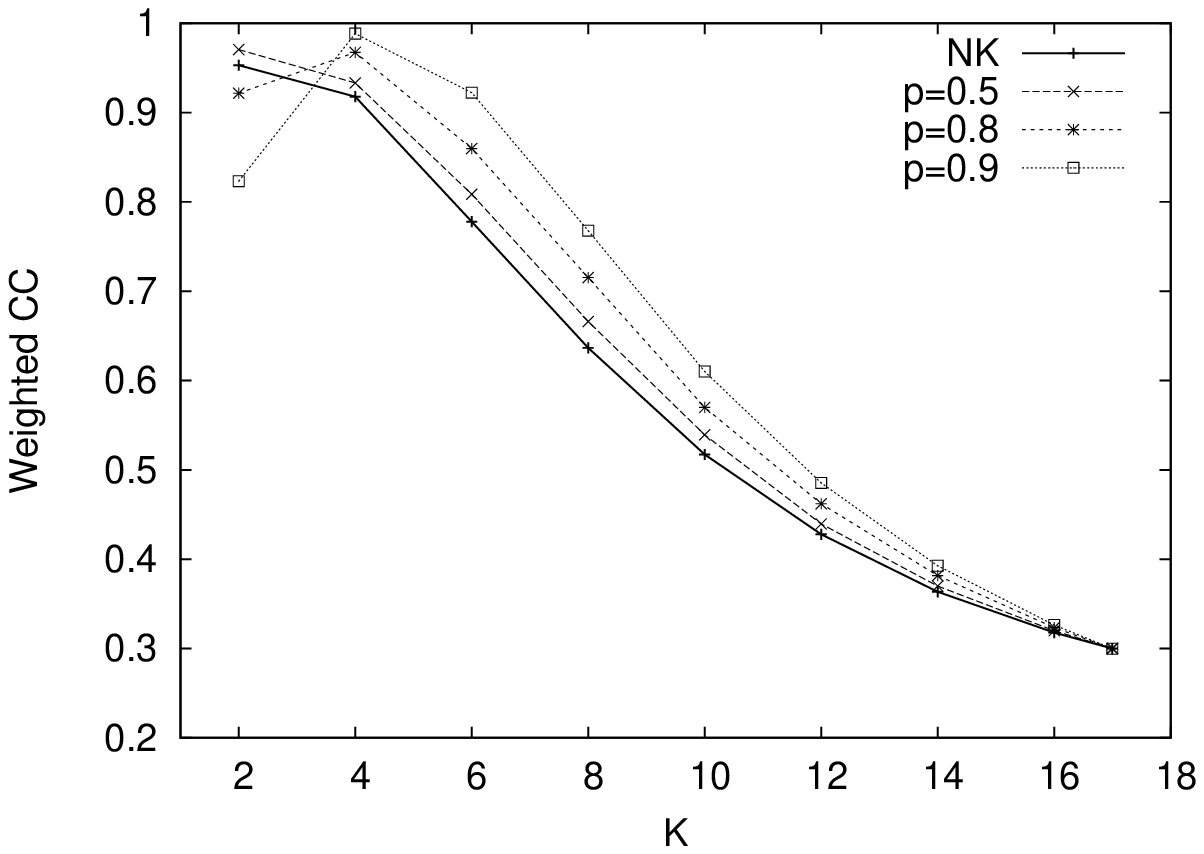} \\
\end{tabular}
    \vspace{-0.3cm}
\caption{Average (30 independent landscapes) of weighted clustering coefficient. $NK_q$ landscapes (top) and $NK_p$ landscapes (bottom).\label{fig:wcc}}
\end{center}
\end{figure}

Figure \ref{fig:wcc} shows the average values of the weighted clustering coefficients for all the combinations of landscape parameters. On both the $NK_q$ and $NK_p$ landscapes, the coefficient decreases with the degree of epistasis and increases with the degree of neutrality.  The decrease in the clustering coefficients with increasing epistasis is consistent with our previous results on standard NK-landscapes \cite{alife08}. For high epistasis and low neutrality, there are fewer transitions between adjacent basins, and/or the transitions are less likely to occur.

\subsubsection{Disparity}

The {\em disparity} measure proposed in \cite{bart05},  $Y_{2}(i)$,  gauges the heterogeneity of the contributions of the edges of node $i$ to the total weight (strength):

$$Y_{2}(i) = \sum_{j \not= i} \left( \frac{w_{ij}}{s_i} \right)^2 $$

\begin{figure} [!ht]
\begin{center}
\begin{tabular}{c}
  \includegraphics[width=0.4\textwidth]{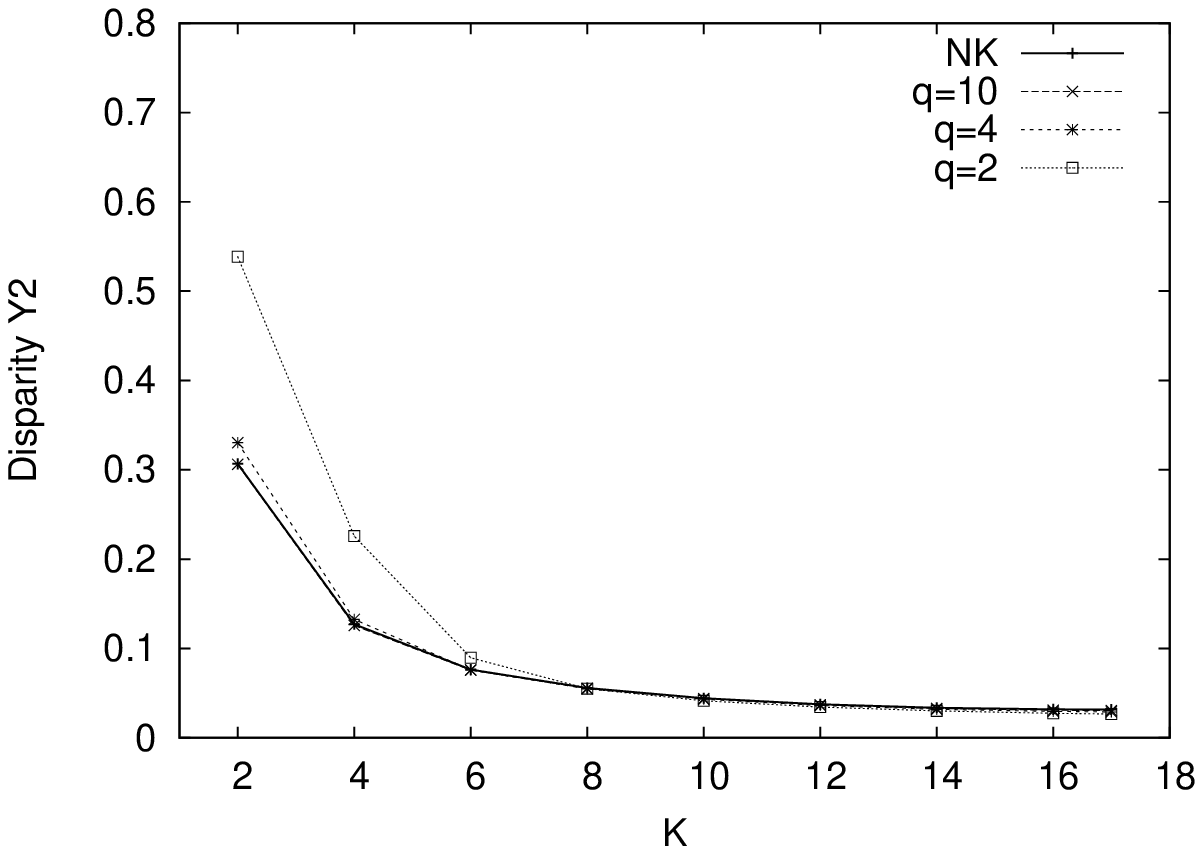}\\
  \includegraphics[width=0.4\textwidth]{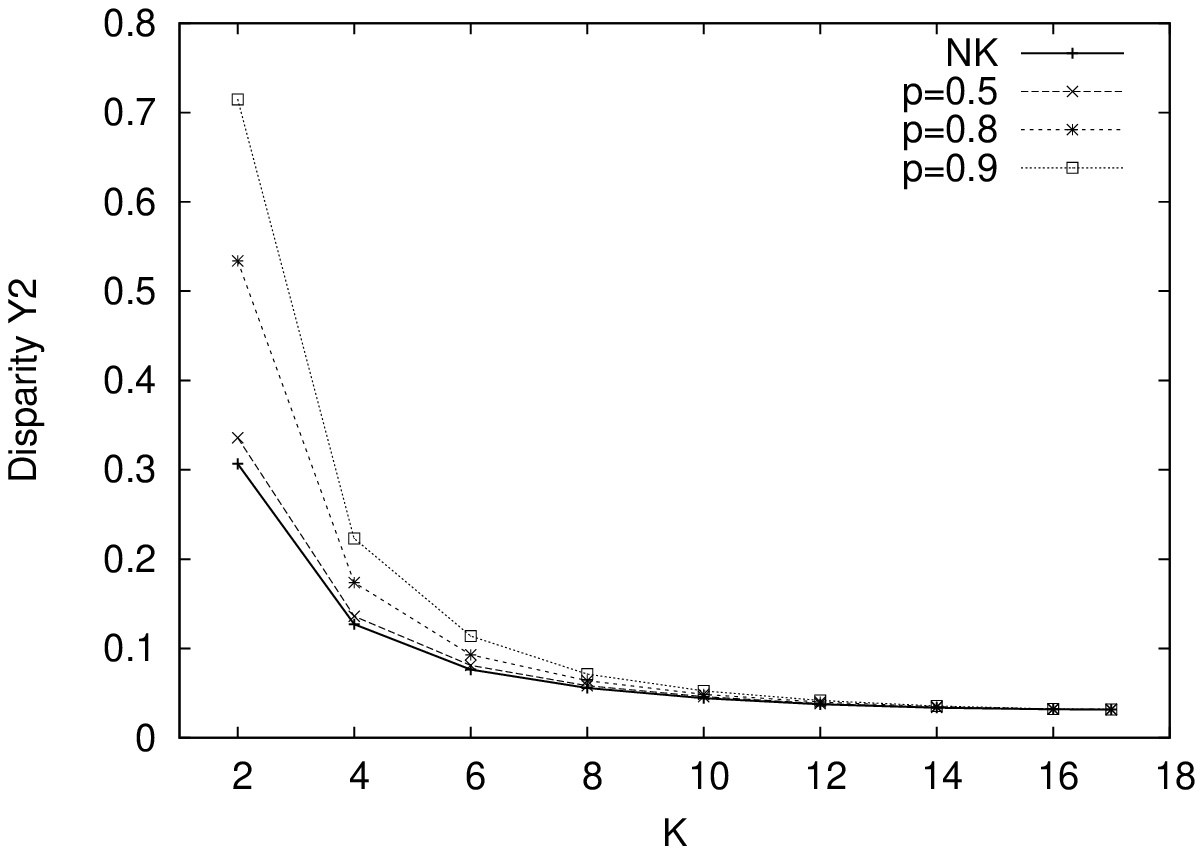}\\
\end{tabular}
\vspace{-0.3cm} \caption{Average disparity $Y_{2}$ for
$NK_q$ landscapes (top) and $NK_p$ landscapes (bottom).
Averages of $30$ independent landscapes.\label{fig:disparity}}
\end{center}
\end{figure}

Figure ~\ref{fig:disparity} depicts the disparity coefficients as defined above. Again the measures are consistent with our previous study on standard $NK$ landscapes \cite{alife08}. Some interesting results with regards to neutrality can also be observed. For low values of $K$, a high degree of neutrality increases the average disparity. When epistasis is high and regardless of the neutrality degree,
the basins are more uniformly connected, and therefore we can picture the local optima network as more "random" \textit{i.e.} more uniform, which has implications on the search difficulty of the underlying landscape.

\subsubsection{Shortest Path}

Finally, as in \cite{alife08,pre09}, in order to compute the shortest distance  between two nodes on the local optima network of a given landscape, we considered the expected number of bit-flip mutations to go from one basin to the other. This expected number can be computed by considering the inverse of the transition probabilities between basins (defined in \ref{defs}). In other words, if we attach to the edges the inverse of the transition probabilities, this value would represent the average number of random mutations to pass from one basin to another.
More formally, the distance between two nodes is defined by $d_{ij} = 1 / w_{ij}$ where $w_{ij}
= p(b_i \rightarrow b_j)$. Now, the length of a path between two nodes is defined as being the sum of these distances along the edges that connect the respective basins. The \textit{average path length} of the whole network is the average value
of all the possible shortest paths.

\begin{figure} [!ht]
\begin{center}
\begin{tabular}{c}
\includegraphics[width=0.4\textwidth]{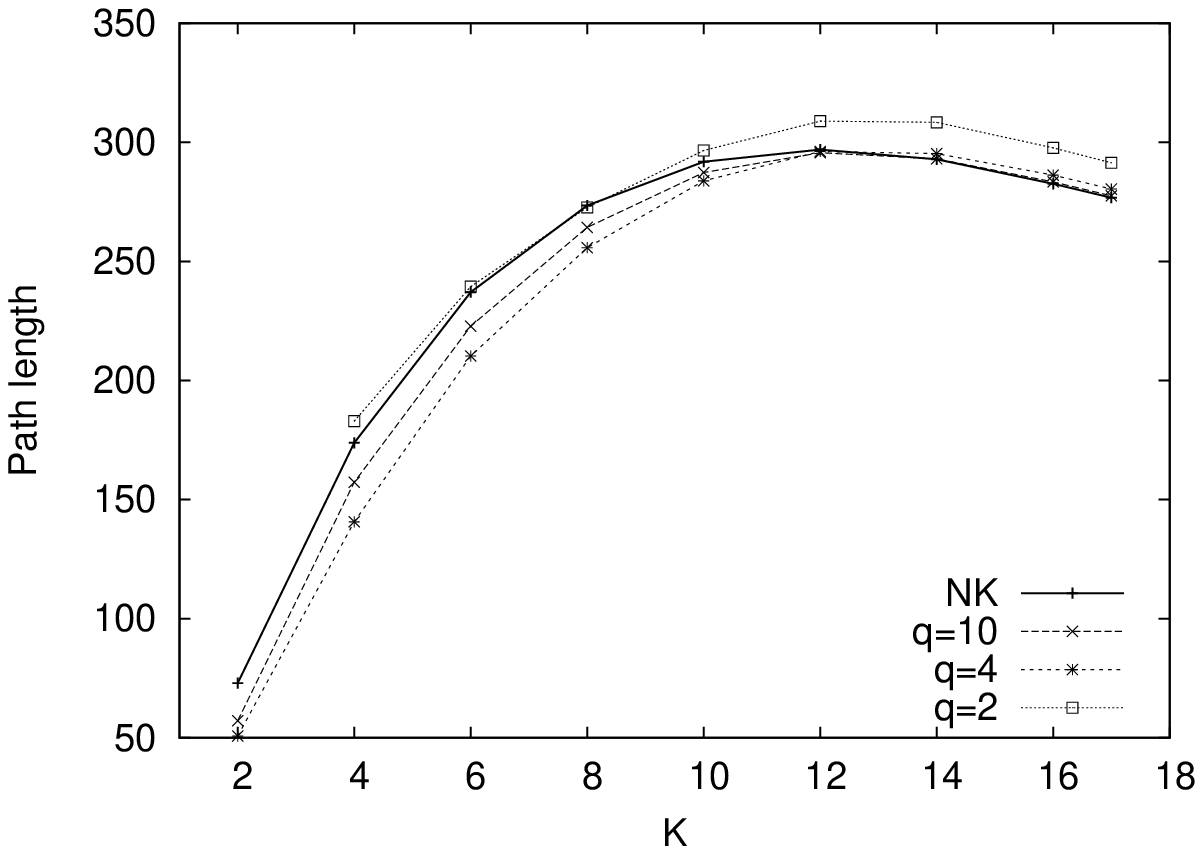} \\
\includegraphics[width=0.4\textwidth]{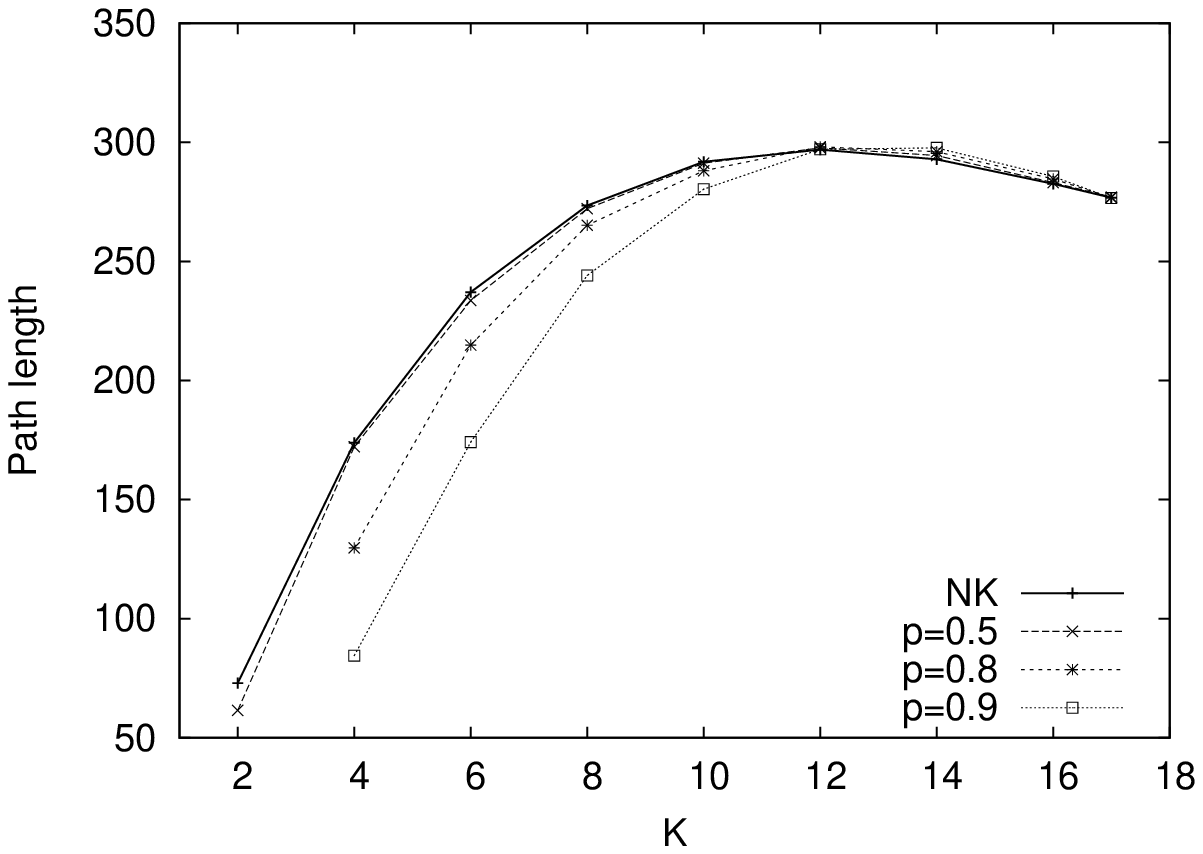} \\
\end{tabular}
    \vspace{-0.3cm}
\caption{Average shortest path lengths between local optima
for $NK_q$ landscapes (top), and $NK_p$ landscapes (bottom).
Averages of 30 independent landscapes.
\label{fig:distances}}
\end{center}
\end{figure}

\begin{figure} [!ht]
\begin{center}
\begin{tabular}{c}
\includegraphics[width=0.4\textwidth]{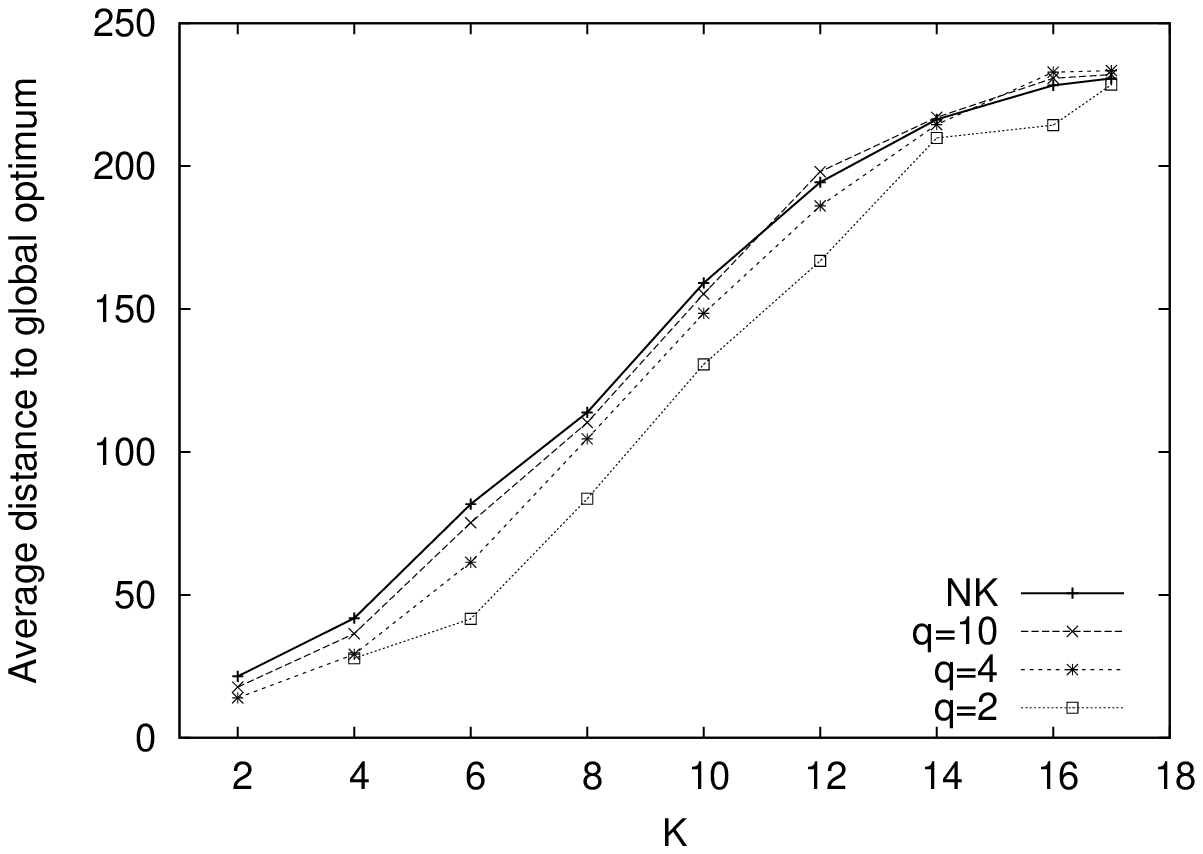} \\
\includegraphics[width=0.4\textwidth]{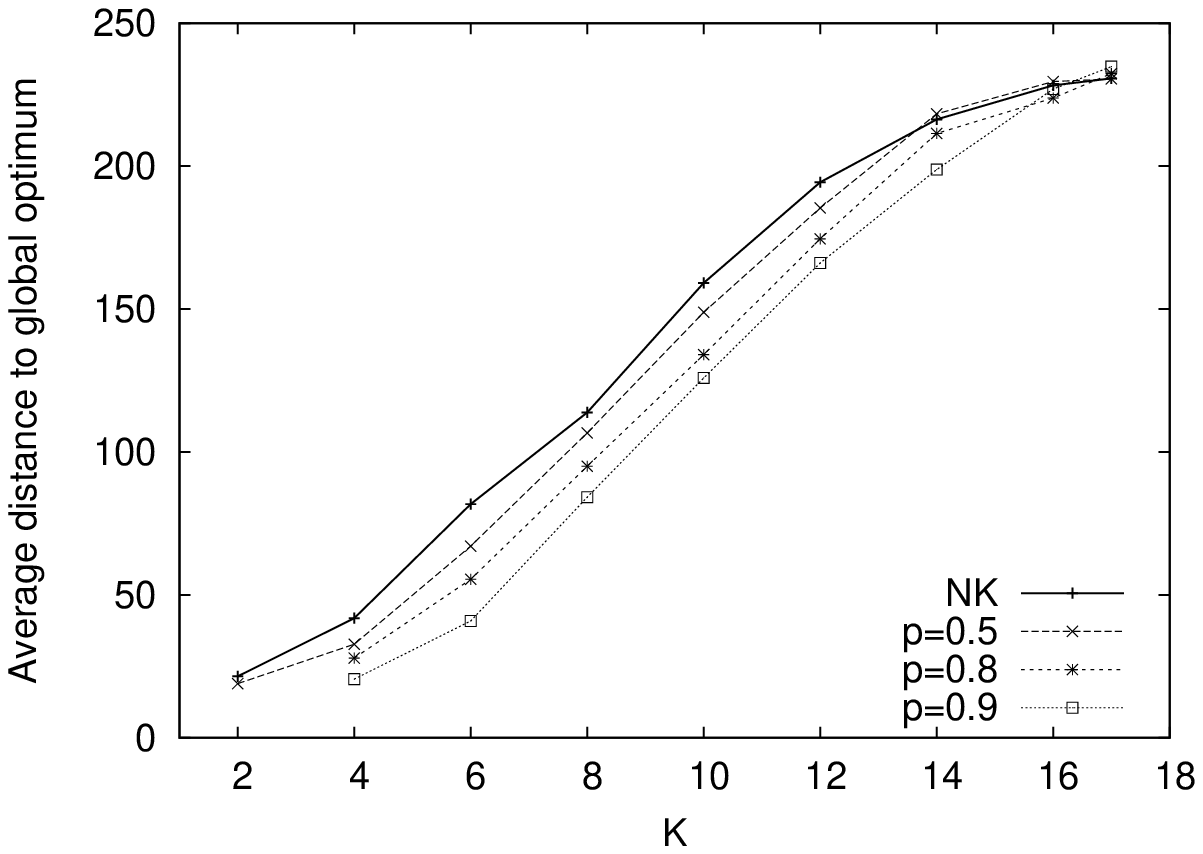} \\
\end{tabular}
    \vspace{-0.3cm}
\caption{Average path length to the optimum from all the other basins
for $NK_q$ landscapes (top), and $NK_p$ landscapes (bottom).
Averages of 30 independent landscapes.
\label{fig:distances_best}}
\end{center}
\end{figure}

Fig.~\ref{fig:distances} is a graphical illustration of the average shortest path length
between basins for all the neutral landscapes studied. The epistasis has the same influence on the results whatever the fa\-mi\-ly of landscapes and the level of neutrality. This path length increases until $K=12$ and decreases thereafter. However, the degree of neutrality introduces some differences between the fa\-mi\-lies; whereas more neutrality decreases the shortest path length for the $NK_p$ family (bottom plot, Fig.~\ref{fig:distances}); the minimal path length is obtained for the intermediate neutrality degrees $q=4$ for $NK_q$ family (top plot, Fig.~\ref{fig:distances}). The longest path length, in this case, is obtained for the largest degree of neutrality ($q=2$). So, even though neutrality is high, the basins are more distant. This confirms that there are structural differences on the two types of landscapes that include neutrality, and some of these structural differences are captured by the local optima networks.

Some paths are more relevant than others from the point of view of a stochastic local search algorithm following a trajectory over the local optima network. In order to better illustrate the relationship of this network property with the search difficulty by heuristic
methods such as stochastic local search, Fig.~\ref{fig:distances_best} shows the shortest path length to the global optimum from all the other basins in the landscape. The trend is clear, the path lengths to the optimum increase steadily with increasing $K$ in all cases. With regards to neutrality, in both types of neutral landscapes, the higher the degree of neutrality, the shortest the path length to the global optimum. This suggest, therefore, that the kind of neutrality introduced in the $NK_p$ and $NK_q$ landscapes could be a positive factor in the search of the global optimum\footnote{The empirical evaluation of search difficulty in $NK_p$ and $NK_q$ landscapes for a standard EA
is studied in the Appendix. It shows that the landscapes with more neutrality (search space size and parameters $K$ being equal) are easier to solve for the EA.}.

\section{Discussion}
\label{discussion}

The fitness landscape concept has proved extremely useful in many fields, and it is especially valuable for the description of the configuration spaces generated by difficult combinatorial optimization problems. In previous work, we have introduced a network-based model that abstracts many details of the underlying landscape and compresses the landscape information into a graph $G_w$ which we have named the \textit{local optima network}~\cite{alife08,pre09}. The vertices of this weighted oriented graph are the local optima of a given fitness landscape, while the arcs are transition probabilities between optima. The same graph also describes the  basins of attraction in the landscape and the adjacency relationship among them. While previous work dealt with non-neutral landscapes, the present paper treats the case of fitness landscapes where neutrality, i.e. groups of configurations with the same fitness are present. Neutrality is a common feature of many landscapes generated by important combinatorial problems, including real-world problems and it is, thus, fundamental to be able to use the network description also in this case. The most difficult aspect is how to define basins of attraction when there are neutral networks in the landscape and how transitions take place between these basins. Our definition in Sect.~\ref{defs} deals with these issues successfully and it is consistent, both conceptually and mathematically, with the previous definition for non-neutral landscapes.

In order to study the applicability of our methodology, we have used synthetic landscapes where the amount of neutrality can be controlled by a parameter. These landscapes, called $NK_p$ and $NK_q$, are neutral variants of the well known $NK$ family of landscapes. This choice also has the advantage of permitting a comparison between neutral and non-neutral variants of the same family of landscapes. We have measured a set of network and basin properties for these three classes. The general observation is that there is a smooth variation with respect to standard $NK$ landscapes when neutrality is gradually introduced. This outcome was somewhat expected and it confirms that our definitions for neutral landscapes are adequate.

Our analysis of the local optima networks concentrates on the inherent structure of the studied landscapes rather than on the dynamics of a search algorithm on such landscapes. However, our findings, summarized below, support the view  that neutrality may enhance evolutionary search \cite{Huynen1996,Wagner2005,HarveyT96,Nimwegen1997,Barnett:97,barnett98,barnett01,Ebner2001,Igel2003}. The empirical study reported in the Appendix further corroborates this view. As discussed in  \cite{Galvan2006}, there is considerable controversy on whether neutrality helps or hinders evolutionary search. This is so, because many studies emphasize algorithm performance, instead of providing an in-depth investigation of the search dynamics. Moreover, there is not a single definition of neutrality, nor an unified approach of adding redundancy to an encoding \cite{Galvan2006}. Our study, however, concentrates on specific model landscapes which posses fitness correlation and selective neutrality. These model landscapes have been found to resemble the properties of biological RNA-folding landscapes. In particular, they feature neutral networks  which have the ``constant innovation'' property \cite{Huynen1996}. This property raises the possibility that (given enough time) almost any possible fitness value can ultimately be attained by the population. The scenario of a population trapped on a local optima vanishes \cite{barnett98}. The detailed  study by Barnett \cite{Barnett:97,barnett98}, illustrates  the dynamics of a simple evolutionary algorithm on several landscapes featuring neutral networks, and compares it with the dynamics on rugged landscapes without neutrality. The dynamics  on both cases are strikingly different (Figs. 4 and 5 in  \cite{barnett98}). On the non-neutral landscape, the population climbs rapidly up the landscape until it reaches a local optimum, at which higher optima are difficult to reach by mutation; the population is effectively trapped.  In the presence of percolating neutral networks, the scenario of entrapment by local optima is evaded; adaptation is characterized by neutral drift punctuated by transitions to higher fitness networks.

We argue that our results are only relevant to  optimization problems that feature percolating neutral networks with similar statistical properties than those present in the model landscapes studied. It is not possible to directly judge the impact of the results for more realistic optimization problems. Therefore, it is important to analyze  more complex genotype-phenotype mappings in future work. It is worth noticing that massively redundant genotype-phenotype mappings, such as those used in  Cartesian Genetic Programming \cite{Miller2006}, have been found to be beneficial to evolutionary search. The application of the local optima network model  in such scenarios is, therefore, a research direction worth exploring.

Our results, which were at least partly unknown to our knowledge, can be summarized as follows.

The optima networks for neutral $NK$ landscapes are  smaller, in terms of the number of nodes, with respect to standard $NK$. Since the number of maxima (nodes) in the landscape increases with $N$ and $K$, search difficulty in general also increases. But for the same $N$, $K$ pair, the search should be easier in neutral $NK$ landscapes, and the difficulty should decrease with increasing neutrality.

The number of edges in the networks gives the average number of possible transitions between maxima. However, it is more interesting to observe the average probabilities, which can be computed from the empirical distribution of the weights for the outgoing edges. It is seen that neutrality increases the probability that a given local optimal configuration escapes its present basin under the effect of a stochastic local search operator.  This observation supports the idea that a heuristic search algorithm with an adequately set mutation rate  could be more effective when neutrality is present, as the opportunity of  finding a promising  (adaptive) search path is increased \cite{barnett01}.

The statistics on the basins of attraction of the landscapes are particularly interesting. The trend is similar to what has been previously reported by the authors~\cite{alife08,pre09} for the standard $NK$ family, but the size of the basins is larger the higher the degree of neutrality, and it decreases exponentially with increasing $K$. Similarly, and as an important particular case, the size of the global maximum basin decreases exponentially with $K$, and increases with increasing neutrality.

The analysis of the clustering coefficient and the disparity, two useful local features of the optima networks, show that the clustering decreases with the degree of epistasis $K$ while, for a fixed $K$, it tends to increase with increasing locality. This is an indirect topological indication of the fact that maxima are more densely connected in the neutral case, which again confirms the easier heuristic search of the corresponding landscapes. The disparity coefficient, on the other hand, says that for high $K$ the basins tend to be randomly connected, independent of the degree of neutrality, a known result confirmed here from the purely network point of view.

Finally, we have statistically analyzed the average shortest paths between nodes in the maxima networks. This is an important characterization of the landscape which is easy to obtain from our maxima networks. It is relevant because it gives useful indications on the average number of transitions that a stochastic local searcher will do between two maxima. In all cases the path length increases with $K$ up to $K=12$ and then stays almost constant or decreases slightly.  Neutrality decreases the mean path length in the $NK_p$ case, while it increases it for the $NK_q$ family. The same trend is observed for the particular average path length from any maximum to the optimum. This last measure gives a rough approximation of the average number of steps a local searcher would perform in the landscape to reach the optimum from any starting local optimal configuration, if it were ``well-informed'', i.e. if it knew
what would be the average best local optimum hop at each step.

\section{Conclusions}
\label{conclusions}

We have found that the topological observation of the local maxima networks of a given fitness landscapes gives both useful information on the problem difficulty and may suggest improved ways of searching them.

However, although we think that our network methodology is promising as a description of both neutral and non-neutral combinatorial landscapes, several issues must be addressed before it acquires practical usefulness. For example, we have limited ourselves to landscape sizes that can be fully enumerated in reasonable time by using relatively low values of $N$. Of course, this is not going to be possible for bigger spaces. Work is thus ongoing to sample the landscapes in a statistically significant way, a step that will allow us to extend the analysis to more interesting problem instances. Second, we plan to extend the present type of analysis to more significant combinatorial optimization problems such as the TSP, SAT, knapsack problems, and several others in order to better understand the relationships between problem difficulty and topological structure of the corresponding networks. Additionally, the analysis of problems with more complex  genotype-phenotype mappings, would help to further enlighten the role of neutrality in evolutionary search.  A further step would be to incorporate and analyze the dynamic aspects of search heuristics operating on  these landscapes.  The ultimate goal would be to try to improve the design of  stochastic local search heuristics by using the information gathered in the present and future work on the local optima and basin networks of several problem classes.

\section*{Acknowledgements}
Gabriela Ochoa is supported by the Engineering and Physical Sciences Research Council, UK, under grant  EPSRC (EP/D061571/1), in which Prof. Edmund K. Burke is the principal investigator.

\appendix
\section*{Assessing the impact of neutrality on evolutionary search}

\begin{table*} [!ht]
\begin{center}
\small \caption{Evolutionary algorithm component choices and parameter settings.} \label{tab:eaparam}
\begin{tabular}{|l|l|l|}
  \hline
  $Component$ & $Choice$ &  $Parameter $ $value(s)$\\ \hline
  Population & random initialisation & size = 100 \\
  Mutation  & bit-flip  mutation & rates =  $\{0.01/N, 0.1/N, 0.5/N, 1/N, 1.5/N, 2/N\}$ \\
  Recombination  &  1-point crossover & rates =   $\{0.0, 0.2, 0.4, 0.6, 0.8, 1.0\}$\\
  Selection & tournament  & size =  2 \\
  Stopping criteria & fixed number of evaluations & $10\%$ of search space size (26215 evaluations) \\
  Replacement & generational with elitism  &  \\
  \hline
\end{tabular}
\end{center}
\end{table*}

This appendix compares the search performance of a standard evolutionary algorithm (EA) running on $NK$ landscapes of equal size and ruggedness (epistasis) level but with different degrees of neutrality. The goal is to asses whether the presence of neutrality in a landscape would enhance evolutionary search.  Given that the fitness value of the global optimum in a $NK$ landscape  depends on its parameters ($N$, $K$, $p$ or $q$), a comparison based on the average best fitness of a number of EA runs is not possible. Therefore, we resort to the success rate as a performance measure. This is possible on the small landscapes explored here as the global optimum is known after the exhaustive exploration for extracting the optima networks. For our empirical study we chose the same  landscape parameters as those used in the main sections of the article. Namely, $NK$ landscapes with $N=18$ and  $K=\{2, 4, 6, 8, 10, 12, 14, 16, 17\}$  with and without neutrality, with  three levels of (increasing) neutrality: $q=\{10, 4, 2 \}$ and $p=\{0.5, 0.8, 0.9 \}$  for the $NK_q$ and $NK_p$ models, respectively.  Table \ref{tab:eaparam} summarizes the  evolutionary algorithm operator choices and parameter settings employed.

\begin{figure} [!ht]
\begin{center}
\begin{tabular}{c}
\includegraphics[width=0.4\textwidth]{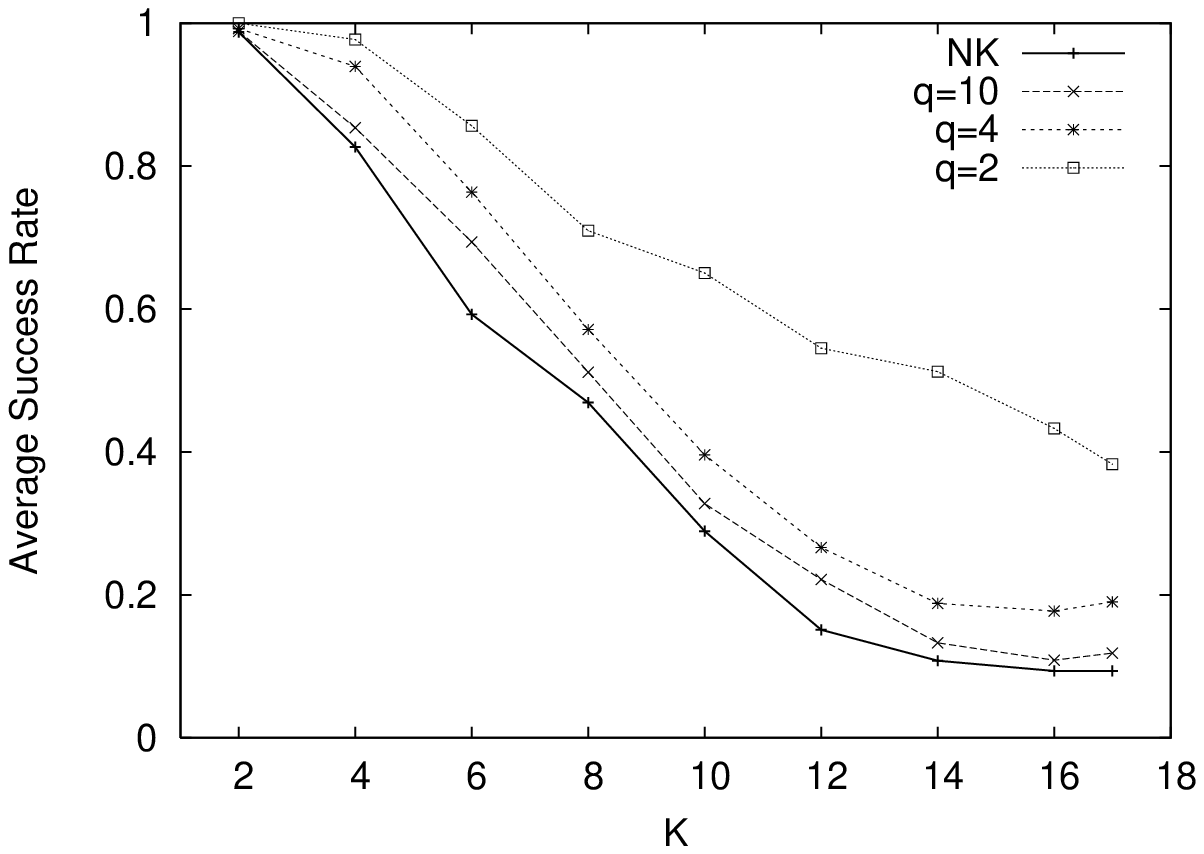} \\
\includegraphics[width=0.4\textwidth]{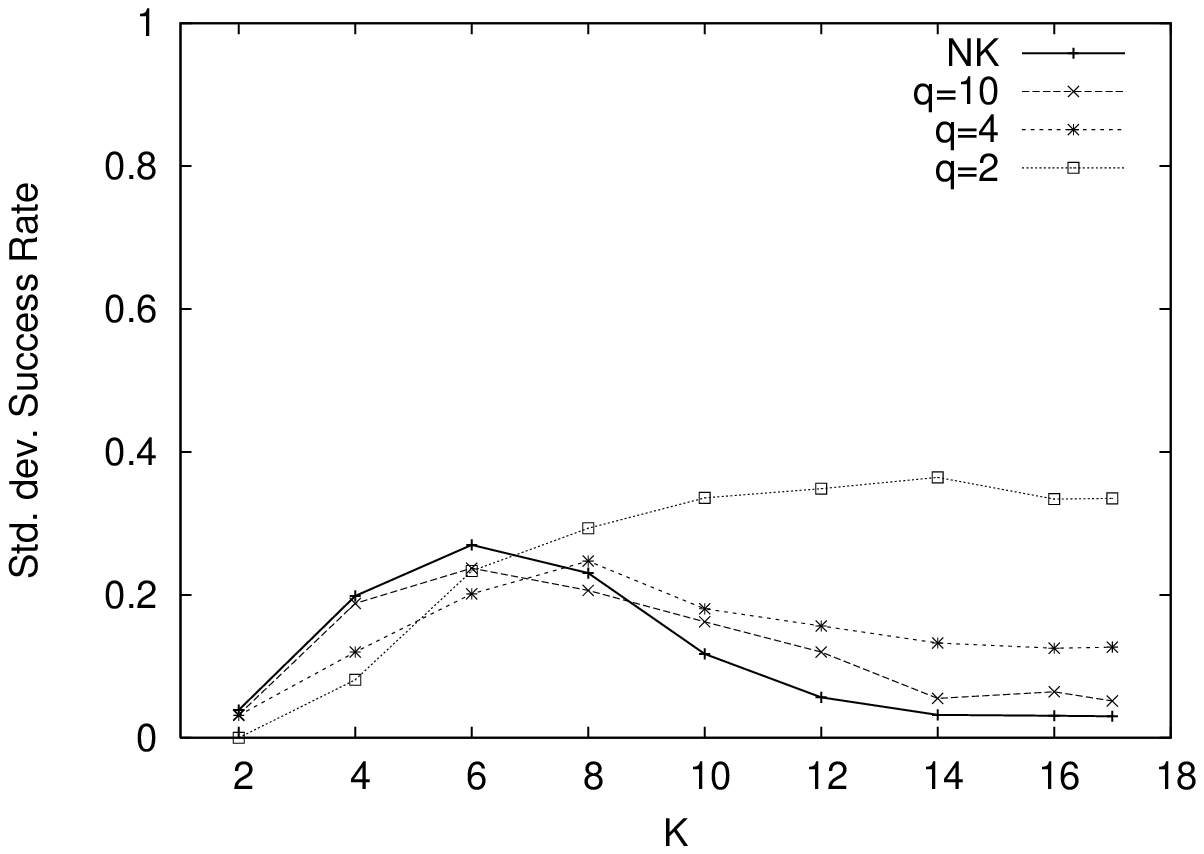} \\
\end{tabular}
\end{center}
\caption{Average (top) and standard deviation (bottom) of the success rate of a standard EA  searching on the $NK_q$ landscapes. See table \ref{tab:eaparam} for EA parameter settings. Averages on 100 independent landscapes.}\label{fig:simpleEANKq}
\end{figure}

\begin{figure} [!ht]
\begin{center}
\begin{tabular}{c}
\includegraphics[width=0.4\textwidth]{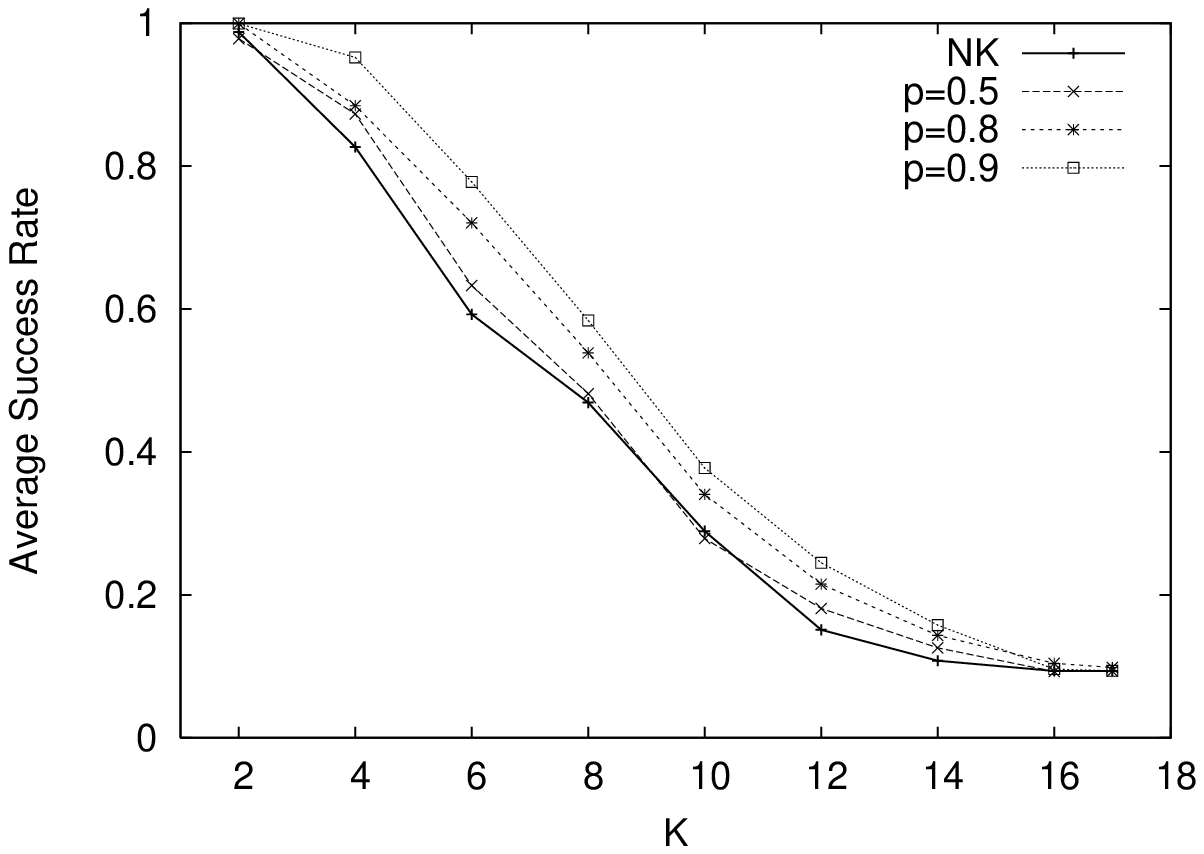} \\
\includegraphics[width=0.4\textwidth]{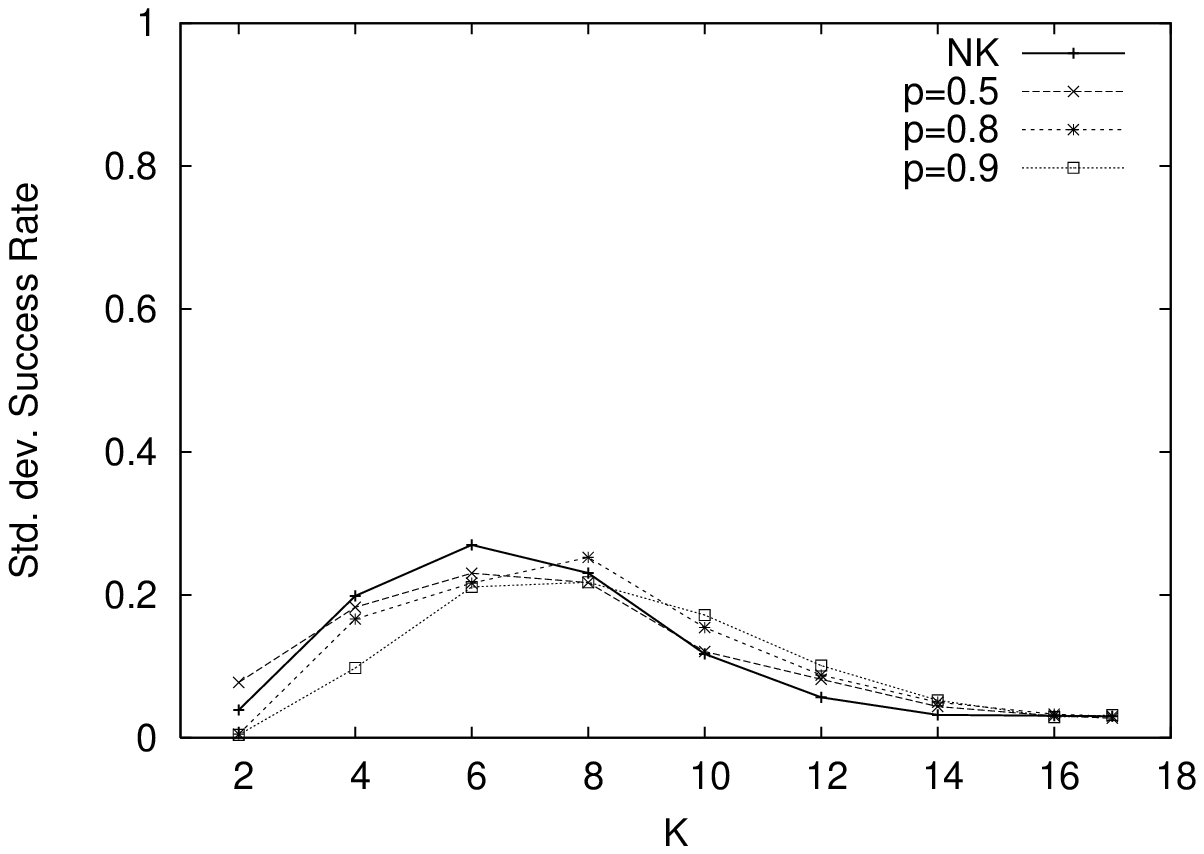} \\
\end{tabular}
\end{center}
\caption{Average (top) and standard deviation (bottom) of the success rate of a standard EA searching on the $NK_p$ landscapes. See table \ref{tab:eaparam} for EA parameter settings. Averages on 100 independent landscapes.}\label{fig:simpleEANKp}
\end{figure}

A preliminary study was carried out to select the optimal combination of mutation and recombination rates for each   $NK$ model and neutrality level. The
 study explored the performance of the $36$ possible mutation and recombination rate pairs (see Table \ref{tab:eaparam}), on 30  independent randomly generated landscape instances of each type. The `optimal' combination was the one achieving the highest average success rate, which is simply defined as the number of runs where the global optimum was found divided by the total number of runs. We found that  the `optimal'  crossover rates were low  (on average $0.1523$ over all landscape types) and the mutation rates per bit were around the well-known figure  $1/N$ \cite{ochoa2006}  (on average $1.317/N$).

To compute the search difficulty on each landscape type, the average and standard deviation of success rates on $100$ runs were computed over $100$ independent landscape instances with the`optimal'  parameter setting found as discussed above.  Figures \ref{fig:simpleEANKq} and  \ref{fig:simpleEANKp}show the average success rates and their standard deviations for the $NK_q$ and the $NK_p$ models, respectively. As it is already  known, the success rates were found to decrease  with increasing epistasis ($K$ values) in all the studied landscapes. Most interestingly, for a given ruggedness level (value of  $K$), the average success rates were found to increase with the degree of  neutrality (figures \ref{fig:simpleEANKq} and  \ref{fig:simpleEANKp}, top plots).  The success rate standard deviations (figures \ref{fig:simpleEANKq} and  \ref{fig:simpleEANKp}, bottom plots) are higher  for $K$ values around 6 except for the $NK_q$ model with $q=2$, for which the standard deviation was found to increase steadily with increasing $K$ values.

\begin{figure} [!ht]
\begin{center}
\begin{tabular}{c}
\includegraphics[width=0.4\textwidth]{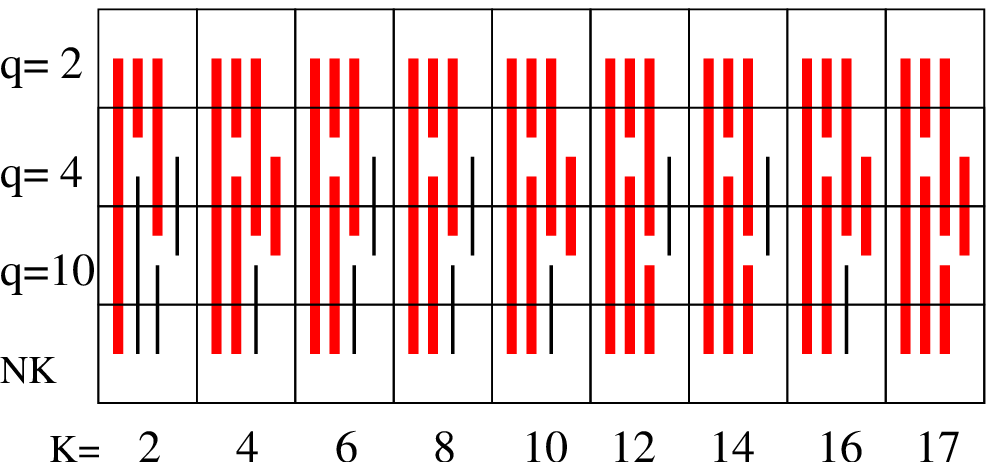} \\
\includegraphics[width=0.4\textwidth]{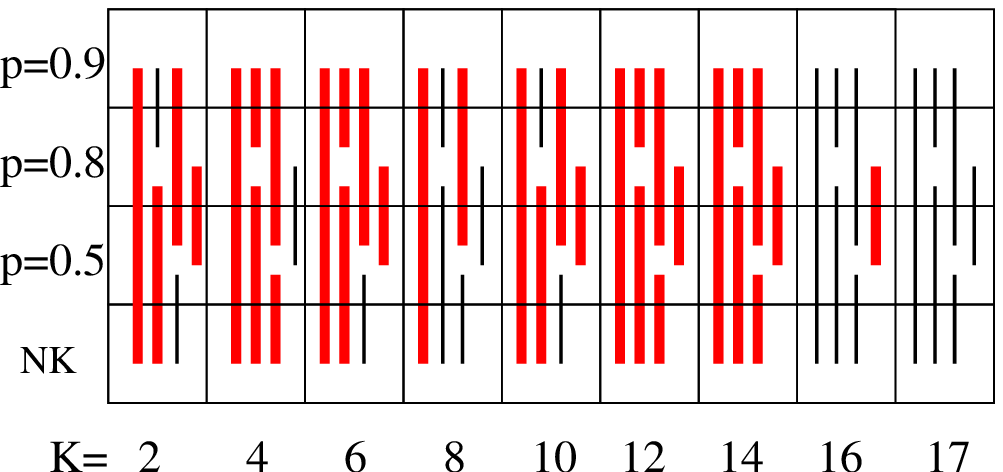} \\
\end{tabular}
    \vspace{-0.1cm}
\caption{Mann-Whitney test to compare the success rate averages of simple EA
on $NK_q$ landscapes (top), and $NK_p$ landscapes (bottom).
A thick line indicates that the equality of average success rates can be rejected with the p-value of $0.05$ according to the test. Otherwise the line is thin.
\label{fig:simpleEAtest}}
\end{center}
\end{figure}

Since the distribution of success rates is not Normal,  we conducted a Mann-Whitney test to asses the statistical significance of  the difference between the averages (see figure \ref{fig:simpleEAtest}). We compared the averages for various  neutrality degrees with the same  epistasis ($K$ value). A thick  line between two neutral parameter values  means that the difference is significant with a p-value of $5\%$; whereas a thin line indicates that the difference between the  averages are not statistically significant. For $NK_q$ landscapes, the average differences are nearly always significant except between some non-neutral $NK$ landscapes and $NK_q$ with low neutrality ($q=10$). Similar results are found for the $NK_p$ model, with the exception  the highest epistasis values where there is nearly no difference between the averages.
Our results clearly suggest that, for the landscape models studied, neutrality increase the evolvability of rugged landscapes. More precisely, $NK$ landscapes of equal size and epistasis level, are  easier to search for a simple EA when  neutrality  is higher.

\end{document}